\documentclass{article}

\ifdefined\pdfminorversion\pdfminorversion=6\fi

\PassOptionsToPackage{numbers, compress}{natbib}
\usepackage[preprint]{neurips_2026}

\usepackage[utf8]{inputenc} 
\usepackage[T1]{fontenc}    
\usepackage{hyperref}       
\usepackage{url}            
\usepackage{booktabs}       
\usepackage{amsfonts}       
\usepackage{nicefrac}       
\usepackage{microtype}      
\usepackage{xcolor}         
\usepackage{graphicx}
\usepackage{subcaption}

\usepackage{amsmath}
\usepackage{amssymb}
\usepackage{mathtools}
\usepackage{amsthm}

\usepackage{multirow}
\usepackage{makecell}
\usepackage{ulem}

\newcommand{\R}{\mathbb{R}}

\title{DeepWaveRL: Self-Supervised Full Waveform Inversion via Reinforcement Learning}

\author{%
  Peng Jin\textsuperscript{1}\quad
  Yinan Feng\textsuperscript{1}\quad
  Yinpeng Chen\textsuperscript{2}\quad
  Youzuo Lin\textsuperscript{1}\quad
  Sharon Huang\textsuperscript{3}\\[0.5em]
  \normalfont\small\textsuperscript{1}University of North Carolina, Chapel Hill, U.S.\\
  \normalfont\small\textsuperscript{2}Google DeepMind, U.S.\\
  \normalfont\small\textsuperscript{3}Pennsylvania State University, U.S.\\[0.5em]
  \normalfont\small\texttt{pjin@unc.edu}, \texttt{ynf@unc.edu}, \texttt{yinpengc@google.com}\\
  \normalfont\small\texttt{yzlin@unc.edu}, \texttt{suh972@psu.edu}
}

\hypersetup{
  pdftitle={DeepWaveRL: Self-Supervised Full Waveform Inversion via Reinforcement Learning},
  pdfauthor={Peng Jin, Yinan Feng, Yinpeng Chen, Youzuo Lin, Sharon Huang}
}

\begin{document}

\maketitle

\begin{abstract}
Full Waveform Inversion (FWI) is a fundamental inverse problem in geophysics to estimate subsurface properties, such as velocity, from seismic measurements. Recently, self-supervised learning methods remove the need for ground-truth velocity maps during training by leveraging a forward operator to simulate seismic data from predictions and backpropagate the gradients of reconstruction loss through it. In this work, we consider a more challenging setting without accessing such gradients. To this end, we incorporate reinforcement learning into self-supervised FWI and propose a framework named DeepWaveRL. Specifically, we reformulate FWI as a policy learning problem, where the policy network predicts velocity maps as actions, and a forward operator is used solely for simulation. A global scalar reward is computed based on data misfit for each prediction and used to weight policy gradients. Thus, learning proceeds without accessing the gradients from the forward operator. Given the indirect and coarse supervision signals, we further identify key strategies to stabilize learning and mitigate the spatial credit assignment bottleneck. Experimental results show that, despite relying entirely on a single scalar reward, DeepWaveRL achieves competitive performance compared to supervised and self-supervised baselines. These findings suggest that our framework offers a flexible and complementary learning paradigm for the FWI research community.
\end{abstract}

\section{Introduction}

\begin{figure*}[t]
    \centering
    \includegraphics[width=0.9\linewidth]{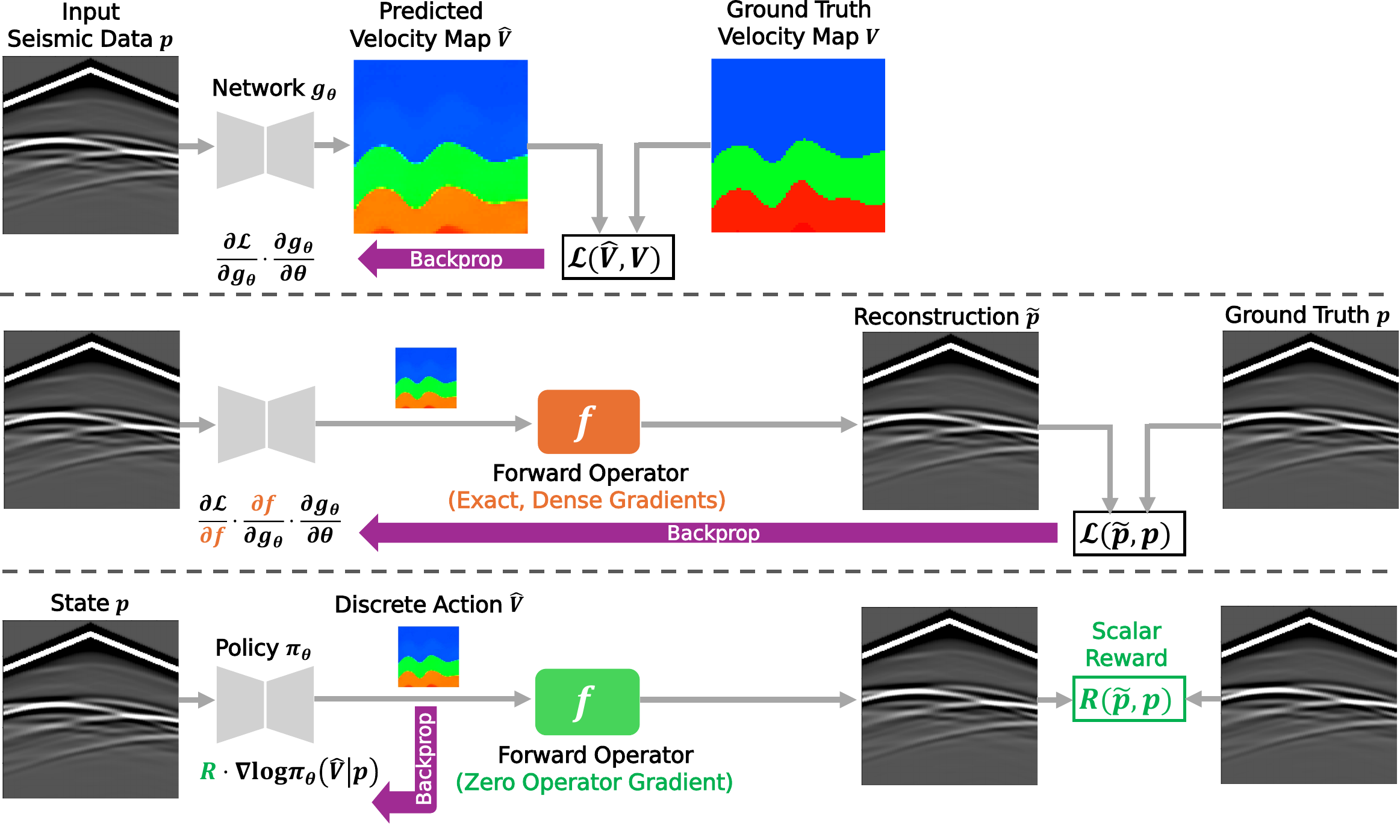}
    \caption{Comparison between data-driven FWI methods. \textbf{Top}: Supervised learning methods compute loss between predicted and ground-truth velocities; \textbf{Middle}: Exact-gradient self-supervised learning method computes the loss between input and reconstructed seismic data and backpropagates gradients through a forward operator $f$; \textbf{Bottom}: Our DeepWaveRL uses the forward operator solely to compute a global scalar reward based on data misfit, without accessing gradients from $f$.}
    \label{fig:teaser}
    \vspace{-2em}
\end{figure*}

Subsurface imaging is essential for characterizing geological structures and geophysical properties (e.g., velocity and impedance), with applications in energy exploration, carbon capture and sequestration, and earthquake early warning systems. A central technique in this domain is Full Waveform Inversion (FWI), which estimates subsurface velocity maps from seismic measurements. Mathematically, for an isotropic medium with constant density, the velocity map and seismic measurements are connected by the acoustic wave equation:
\begin{equation} \label{eq:fwi}
  \nabla^2 p(x,z,t) - \frac{1}{v(x,z)^2} \frac{\partial^2p(x,z,t)}{\partial t^2} = s(x,z,t)\;,
\end{equation}
where $x$ denotes the horizontal offset, $z$ the depth, $p(x,z,t)$ the pressure wavefield at spatial location $(x, z)$ and time $t$, $v(x,z)$ the wave propagation velocity at $(x, z)$ , $s(x,z,t)$ the source term, and $\nabla^2$ the Laplacian operator. In practice, seismic data are often collected at the surface (i.e., $p(x,z=0,t)$). While FWI has the potential to produce high-resolution velocity maps, the inverse problem itself is inherently non-linear and ill-posed. Conventional physics-driven approaches requires intensive computation due to repeated forward simulations per sample and exhibit strong sensitivity to noise and initial conditions. These challenges have motivated growing interest in data-driven FWI methods.

A majority of data-driven methods~\citep{wu2019inversionnet,zhang2019velocitygan, jin2024empirical} adopt the supervised learning paradigm and formulate FWI as an image-to-image translation task. As shown in Figure~\ref{fig:teaser}, deep neural networks directly learns the mapping from seismic data to velocity maps, enabling fast inference and achieving high accuracy. However, such methods depend on large collections of paired seismic data and velocity maps for training. In real-world scenarios, ground-truth velocity maps are rarely available, as constructing them is extremely time-consuming and requires substantial expertise from geophysicists.

Recent work (UPFWI)~\citep{jin2021unsupervised} explicitly leverages the underlying physics equations and eliminates the need for ground-truth velocity maps via self-supervised learning. As illustrated in Figure~\ref{fig:teaser}, UPFWI couples a network with a differentiable forward operator $f$ to simulate seismic data from predictions. Training is driven by minimizing a reconstruction loss in seismic domain, with \textit{gradients backpropagated through $f$}. While this approach enables end-to-end learning without ground-truth velocity maps, it relies on access to exact, dense gradients from the forward operator.

In this paper, we push the boundaries of self-supervised FWI by investigating a more challenging optimization regime: learning the high-dimensional spatial mapping using only \textit{a single global scalar reward} without accessing the gradient information from the forward operator. To this end, we present DeepWaveRL, a self-supervised framework for FWI that incorporates reinforcement learning. As depicted in Figure~\ref{fig:teaser}, we formulate FWI as a single-step decision problem, where the input seismic data serve as the state, and a policy network outputs the velocity map as the action. A forward operator $f$ is used to simulate seismic data from the prediction and to compute a global scalar reward based on the seismic data misfit. Policy optimization is performed using a group-based policy gradient algorithm, which updates the network parameters and action distributions by weighting log-probability gradients with the group-relative advantages. 

Learning under this strict scalar-reward regime poses several challenges. First and foremost is the spatial credit assignment bottleneck. Because the reward evaluates the entire 2D velocity map globally, the agent lacks localized feedback to determine which specific pixels or structures caused the error. Second, the vast spatial action space triggers a combinatorial explosion, leading to extreme variance in the policy gradient estimator. Finally, seismic measurements exhibit large dynamic ranges, where dominant directive energy overwhelms weaker but informative deeper reflections. This leads to a highly non-convex physical reward landscape and makes stable exploration exceptionally difficult.

To stabilize policy gradients and mitigate these challenges, we introduce three targeted algorithmic strategies. First, we employ a map-level advantage estimator combined with a discretized action space, which effectively bounds the exploration manifold and dramatically reduces the variance of the gradient updates. Second, we introduce a sign-preserving logarithmic reward transformation to reshape the physical reward landscape, compressing dominant wave energy to extract learning signals for deep-region structures. Finally, we demonstrate that initializing the policy with learned structural priors by transfer learning acts as a crucial regularizer, safely guiding the agent past local minima.

To evaluate the efficacy of our framework, we benchmark DeepWaveRL on several datasets from OpenFWI~\citep{deng2022openfwi}, a large-scale, multi-structural dataset collection. Experimental results show that, despite relying entirely on a single scalar reward, our DeepWaveRL attains comparable performance to the supervised baseline InversionNet~\citep{wu2019inversionnet,jin2024empirical} on CurveVel-A, with a Mean Absolute Error (MAE) of 0.0527 (vs. 0.0409), a Root Mean Squared Error (RMSE) of 0.1012 (vs. 0.0944), and a Structured Similarity (SSIM) of 0.8601 (vs. 0.8796). Furthermore, we demonstrate that leveraging learned structural priors via transfer learning allows DeepWaveRL to generalize efficiently to complex new fault geometries in FlatFault-A and CurveFault-A.

Our contribution is summarized as follows: 
\begin{itemize}
  \item \textbf{DeepWaveRL Framework:} We introduce DeepWaveRL, a reinforcement learning framework for self-supervised FWI under the strict scalar-reward regime.
  \item \textbf{Stabilization and Spatial Credit Assignment:} We stabilize policy gradients across a massive combinatorial action space and mitigate the spatial credit assignment bottleneck in the high-dimensional domain by integrating a map-level advantage estimator, a discretized spatial action space, and a sign-preserving non-linear reward transformation.
  \item \textbf{Competitive Empirical Performance:} We demonstrate that DeepWaveRL achieves competitive performance using only a scalar data-misfit reward without requiring access to ground-truth velocity maps or gradients from the forward operator.
  \item \textbf{Empirical Variance Analysis:} We provide an empirical analysis detailing the variance-reduction properties of group-based policy optimization in FWI. We explicitly demonstrate how discrete action spaces and prior regularization overcome the optimization issues in high-dimensional continuous environments.
\end{itemize}

\section{Method}
In this section, we first briefly summarize the state-of-the-art group-based reinforcement learning algorithms and then present our DeepWaveRL and its components. After that, we provide the comparison of DeepWaveRL with other data-driven methods from a gradient perspective.

\subsection{Preliminary}
\citet{deepseek-math} introduces Group Relative Policy Optimization (GRPO) that enhances Proximal Policy Optimization (PPO)~\citep{schulman2017proximal} by omitting the value function and estimating the advantage in a group-relative manner. This is followed by several variants such as Decoupled Clip and Dynamic Sampling Policy Optimization (DAPO)~\citep{yu2025dapo} and Group Sequence Policy Optimization (GSPO)~\citep{zheng2025group}, yielding superior training efficiency and performance. The core idea of GRPO is as follows.

For a specific question $q\sim P(Q)$, a group of $G$ responses $\{ o_i\}_{i=1}^G$ are sampled from an old policy network $\pi_{\theta_\text{old}}$. Each response $o_i$ is then fed into a reward function to obtain the individual reward $R_i$. By normalizing the rewards within each group, an advantage $\hat{A}_i$ is assigned to each response. The policy network is optimized by maximizing the following clipped objective, similar to PPO:
\small
\begin{equation}
\begin{aligned}
\mathcal{J}_{\text{GRPO}}&(\theta) = \mathbb{E}_{{q\sim P(Q), \{o_i\}_{i=1}^G\sim \pi_{\theta_\text{old}}(\cdot \mid q)}} \frac{1}{G}\sum_{i=1}^{G}\frac{1}{|o_i|}\sum_{t=1}^{|o_i|} \Big[ 
L(r_{i,t}(\theta), \hat{A}_{i,t}, \epsilon) - \beta D_{\text{KL}}(\pi_{\theta} || \pi_{\text{ref}}) 
\Big], 
\label{eq:grpoloss}
\end{aligned}
\end{equation}
and the $L$ is given by 
{
\begin{equation}
\begin{aligned}
L &= \min \Big[ r_{i,t}(\theta) \hat{A}_{i,t},
\text{clip} \Big( r_{i,t}(\theta), 1 - \varepsilon, 1 + \varepsilon \Big) \hat{A}_{i,t} \Big].
\end{aligned}
\end{equation}
}

Here, $\epsilon$ is the hyperparameter to determine the clipping boundaries, and $\beta$ is to control the importance of the KL divergence $D_\text{KL}$ between the online policy $\pi_{\theta}$ and the frozen reference policy $\pi_{\text{ref}}$. In addition, $\hat{A}_{i, t}$ and $r_{i,t}(\theta)$ are the group-based advantage estimation and importance ratio of $o_{i,t}$, which is the $t-$th token in response $o_i$. They are defined as: 
\begin{equation}
\label{eq:adv_ratio}
\hat{A_{i,t}} = \hat{A_{i}} = \frac{R_i - \text{mean}(\{R_i\}_{i=1}^G)}{\text{std}(\{R_i\}_{i=1}^G)}, \qquad
r_{i,t}(\theta) = \frac{\pi_{\theta}(o_{i,t} \mid q, o_{i,<t})}{\pi_{\theta_{\text{old}}}(o_{i,t} \mid q,o_{i,<t})}.
\end{equation}

\begin{figure*}[t]
    \centering    \includegraphics[width=1.0\linewidth]{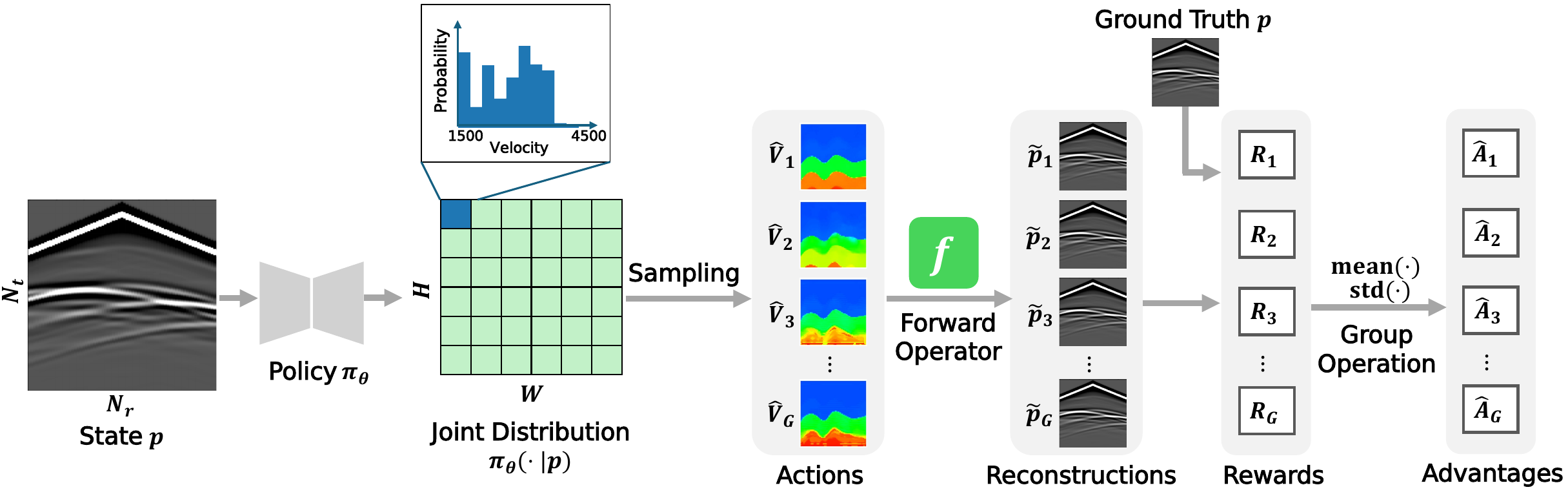}
    \caption{Schematic illustration of the policy optimization pipeline in DeepWaveRL. For brevity, we only show the seismic data from one source.}
    \label{fig:policy}
    \vspace{-0.5em}
\end{figure*}
\subsection{Self-Supervised FWI via Reinforcement Learning}
As illustrated in Figure~\ref{fig:policy}, DeepWaveRL formulates FWI as a single-step policy learning problem. Each input seismic data $p\in \R^{N_s\times N_t\times N_r}$ is treated as a state and passed into the policy network $\pi_\theta$. Here, $N_s$ denotes the number of sources used during data acquisition, $N_t$ the number of recorded timesteps, and $N_r$ the total number of receivers. The network then outputs the probability distribution of velocity $\pi_\theta(\cdot|p)_{h,w}$ at each spatial location $(h, w)$. Following GRPO, we sample a group of $G$ velocity maps $\{\hat{V_i}\}_{i=1}^{G}$ from the predicted joint probability distribution as 2D actions, where each $\hat{V_i}=\{\hat{v}_{i,h,w}\}_{h=1}^{H},_{w=1}^{W}$, and $H$ and $W$ are the vertical and horizontal dimensions of the velocity map. To reduce exploration burden, we use a discrete action space instead of the continuous one. 

To assign a reward $R_i$ to each sampled action $\hat{V_i}$, we employ a forward operator $f$ to simulate seismic data $\tilde{p_i}=f(\hat{V_i})\in \R^{N_s\times N_t\times N_r}$. The reward is then computed based on the misfit between the input seismic data $p$ and the reconstruction $\tilde{p_i}$. We further compute the relative advantage $\hat{A_i}$ within each group following Equation~\ref{eq:adv_ratio}, but use a map-level importance ratio. 

According to a recent work~\citep{zheng2025group}, the mismatch between the unit of reward and the unit of optimization objective can introduce high-variance noise and further lead to model collapse. In our settings, a reward is assigned to the whole 2D velocity map. Therefore, instead of computing the pixel-level importance ratios in Equation~\ref{eq:adv_ratio}, we define the map-level importance ratio as:
\begin{equation}
\begin{aligned}
    m_{i} &= \Bigg[\frac{\pi_\theta(\hat{V}_i|p)}{\pi_{\theta_\text{old}}(\hat{V}_i|p)}\Bigg]^{\frac{1}{H\cdot W}} = \text{exp} \Bigg[\frac{1}{H\cdot W} \sum_{h=1}^{H} \sum_{w=1}^{W} \text{log}(\frac{\pi_\theta(\hat{v}_{i,h,w}|p)}{\pi_{\theta_\text{old}}(\hat{v}_{i,h,w}|p)}\Bigg].
\end{aligned}
\end{equation}

Consequently, the overall optimization objective can now be written as:
\small
\begin{equation}
\begin{aligned}
\mathcal{J}&(\theta) = \mathbb{E}_{{p\sim \mathcal{P}, \{\hat{V}\}_{i=1}^G\sim \pi_{\theta_\text{old}}(\cdot \mid p)}} \frac{1}{G}\sum_{i=1}^{G} \Big\{ 
\min \Big[ m_{i}(\theta) \hat{A}_i, \text{clip} \Big( m_{i}(\theta), 1 - \varepsilon_\text{low}, 1 + \varepsilon_\text{high} \Big) \hat{A}_i \Big] \Big\},
\label{eq:loss}
\end{aligned}
\end{equation}
where we follow \citet{yu2025dapo} and decouple the lower and higher clipping range as $\varepsilon_\text{low}$ and $\varepsilon_\text{high}$, and $\mathcal{P}$ denotes the distribution of seismic data. We also remove the KL penalty term as our initial model is not as good as common pretrained language models, and we allow the model distribution to diverge from the initial model.

The policy network is thus trained to shift its output distribution toward higher-reward actions. Importantly, the entire training process is self-supervised, without the involvement of ground-truth velocity maps. Noteworthy, the optimization is driven entirely by the global scalar reward, completely removing the reliance on localized, pixel-level gradients from the forward operator. 

Additionally, because this scalar-reward formulation requires only the input seismic data $p$ and the forward operator to compute rewards, the policy network can undergo Test-Time Optimization (TTO). During inference, TTO fine-tunes the network weights on unseen test samples using the exact same policy optimization objective (Equation~\ref{eq:loss}). This allows the agent to leverage its learned global structural prior as a stable initialization for sample-specific iterative refinement. This test-time optimization further boosts performance, which will be discussed in Section~\ref{sec:main_results}.

\subsection{Strategies for Variance Reduction, Credit Assignment and Regularization}
\label{sec:strategies}
\textbf{Discrete Action Space.}
During training, we treat the predicted velocity map $\hat{V}=\{\hat{v}_{h,w}\}_{h=1}^{H},_{w=1}^{W}$ as a 2D action sampled from the joint distribution. We initially experimented with a continuous action space, where the network outputs two maps of size $H\times W$, representing the mean $\mu_{h,w}$ and standard deviation $\sigma_{h,w}$ of a Gaussian distribution at each spatial location. Hence, the velocity at location $(h, w)$ was sampled as $\hat{v}_{h,w} \sim \mathcal{N}(\mu_{h,w}, \sigma_{h,w})$. However, this formulation led to highly unstable training and noisy predictions due to the extreme variance of the policy gradient estimator in a high-dimensional continuous space. To bound the exploration manifold and reduce this variance, we discretize the velocity values uniformly into $B$ bins, treating the sampled action as a predicted category. Given an action $a_{h,w}\in [0,1,...,B-1)$, we compute the corresponding velocity as
\begin{equation}
    v(a_{h,w})=\frac{v_\text{max} - v_\text{min}}{B} \cdot (a_{h,w} +0.5) + v_\text{min},
\end{equation}
where $v_\text{max}$ and $v_\text{min}$ are the largest and smallest possible velocities in a dataset. This discrete action space substantially improves both training stability and prediction quality by regularizing the combinatorial search space. 

\textbf{Sign-Preserving Logarithm Transformation in Reward.}
Similar to the loss function of UPFWI~\citep{jin2021unsupervised}, we define the reward as the negative pixel-wise $\ell_1$ and $\ell_2$ distance between the input and reconstructed seismic data. To further enhance learning in the scalar-reward regime, we apply a sign-preserving logarithm transformation during the computation of rewards as:
\begin{equation}
    p' = \text{sign}(p) \cdot \text{log}(k\cdot |p| + c),
\end{equation}
where $k$ and $c$ are hyperparameters to control the strength of the transformation. This non-linear transformation can compress dominant directive wave energy and amplify weaker signals (e.g., reflections and deep arrivals). Without the transformation, the advantage estimator assigns effectively zero credit to improvements in deep-region structures. By balancing the spatial sensitivity of the scalar reward, the log transformation ensures that structural updates in deeper regions yield meaningful, localized gradient signals. The reward function can then be described as:
\begin{equation}
    R = -\ell_1(p', \tilde{p}') - \ell_2(p', \tilde{p}').
\end{equation}

\textbf{Transfer Learning using Well-Trained Policies.}
Directly training our DeepWaveRL from scratch sometimes leads to unstable learning and convergence issues. We provide examples of predicted velocity maps generated by these models in Appendix~\ref{fig:unstable}. The predictions exhibit unrealistically low velocities in deep regions after certain training steps and cannot recover. This is a direct consequence of the agent encountering local minima in the highly non-convex reward landscape. To address this issue, we propose to initialize the policy network with weights from a well-trained model on a different dataset. This transfer learning strategy allows prior knowledge of spatial velocity distributions to be reused across datasets, acting as a strong structural prior that regularizes the agent's exploration to physically plausible manifolds. This eventually leads to more stable training and improved convergence.

\subsection{Comparison with Previous FWI Methods from a Gradient Perspective}

To demonstrate the relationship among supervised, exact-gradient self-supervised, and our scalar-reward RL-based self-supervised approach, we provide analysis from the perspective of gradient construction and propagation. Full derivations and detailed comparisons are provided in Appendix~\ref{sec:gradient_comparison}.

\textbf{Supervised:} Gradients are directly computed from discrepancies with ground-truth velocity maps, $\nabla_\theta \mathcal{L}_{\mathrm{sup}} \propto \frac{\partial \,\|\, V - \hat{V} \,\|}{\partial \hat{V}}$. While this yields stable optimization and strong supervision, it is impractical in real-world scenarios due to the scarcity of paired data.

\textbf{Exact-Gradient Self-Supervised:} Methods like UPFWI remove the need for ground-truth maps by backpropagating exact, dense gradients through a fully differentiable forward operator $f$, yielding $\nabla_\theta \mathcal{L}_{\mathrm{diff}} \propto  \frac{\partial f(\hat{V})}{\partial \hat{V}}$. While this enables physics-guided learning, it fundamentally relies on access to analytical gradients, coupling the optimization tightly to $f$.

\textbf{Scalar-Reward RL-Based self-supervised:} In contrast to exact-gradient methods, DeepWaveRL converts the global reconstruction error into a scalar reward signal that acts merely as a multiplier for the policy gradients: $\nabla_\theta \mathcal{J}(\theta) \propto \mathbb{E}\left[ \hat{A} \cdot \; \nabla_\theta \log m_i(\theta) \right]$.

\section{Experiments}
In this section, we evaluate the performance of DeepWaveRL on OpenFWI~\citep{deng2022openfwi}. We also investigate the impact of the discrete action space and other variance reduction components through ablation studies.

\subsection{Datasets}
We verify our method on CurveVel-A (CVA), FlatFault-A (FFA) and CurveFault-A (CFA) of OpenFWI~\citep{deng2022openfwi}, an open-source collection of large-scale, multi-structural benchmark datasets for data-driven seismic FWI.

CVA contains velocity maps composed of curved layers with clear interfaces, while FFA and CFA focus more on geological fault identification and have velocity maps with flat and curved layers, respectively. CVA contains 30K pairs of seismic data and velocity maps. Following the official data split, we use 24K for training and 6K for testing. FFA and CFA contain 54K samples each, and we use 48K/6K splitting.

Each velocity map in all three datasets has a size of $70\times70$, with a grid size of 10 meters in horizontal and depth directions. The velocity value ranges from 1,500 meter/second to 4,500 meter/second. For seismic data, five sources are placed evenly with a 170-meter spacing and a central source frequency of 15 Hz. The seismic data are recorded by 70 receivers at 10-meter intervals, each collecting 1,000 timesteps over 1 second. This results in seismic data of shape $5\times1000\times70$. For additional details, we refer readers to the original OpenFWI paper~\citep{deng2022openfwi}.


\subsection{Experiment Settings}
\textbf{Evaluation Metrics:} We evaluate predicted velocity maps using MAE, RMSE, and Structural Similarity (SSIM), consistent with prior work~\citep{wu2019inversionnet, feng2024auto, deng2022openfwi}. MAE and RMSE quantify pixel-wise errors, while SSIM captures perceptual similarity, reflecting the structured information of velocity maps where distortions can be easily perceived by a human. Note that all measurements are computed on normalized velocity maps, with MAE and RMSE in the range $\left[-1, 1\right]$, and SSIM in $\left[0, 1\right]$.

\textbf{Comparison:} We compare our method to InversionNet~\citep{wu2019inversionnet} which achieves the state-of-the-art performance when trained solely on each dataset~\citep{jin2024empirical}. Following the configurations in OpenFWI, we train UPFWI~\citep{jin2021unsupervised} as a self-supervised baseline. Additionally, we implement simultaneous perturbation stochastic approximation (SPSA)~\citep{spall2002multivariate} as a derivative-free optimization baseline. However, its performance is not comparable to other methods, so we report the results and details of SPSA in Appendix~\ref{sec:zero}.

\textbf{Implementation:} We follow the UPFWI paper and incorporate a finite-difference forward operator. For DeepWaveRL, as no gradient from the forward operator is required, we develop a custom, pre-compiled CUDA operator that includes in-place operations for acceleration. Training details are provided in Appendix~\ref{sec:tech}.

\begin{table*}[t]
\caption{Quantitative results evaluated on CVA. For models with discrete predictions, we report the mean estimate.}
\label{tab:main}
\begin{center}
\setlength{\tabcolsep}{20pt}
\renewcommand{\arraystretch}{1.1}
\vspace{-1em}
\small
\begin{tabular}{c|c|c|c}
\Xhline{4\arrayrulewidth}
Method & MAE$\downarrow$& RMSE$\downarrow$ & SSIM$\uparrow$ \\
\Xhline{4\arrayrulewidth}
InversionNet & \textbf{0.0409} & \textbf{0.0944} & \textbf{0.8796} \\ \hline
UPFWI & 0.0739 & 0.1369 & 0.7420 \\ \hline
DeepWaveRL & 0.0717 & 0.1300 & 0.8303 \\ \hline
DeepWaveRL + TTO & \uwave{0.0527} & \uwave{0.1012} & \uwave{0.8601} \\
\Xhline{4\arrayrulewidth}
\end{tabular}
\end{center}
\vspace{-1.5em}
\end{table*}

\begin{figure*}[t]
    \centering
    \includegraphics[width=1.0\linewidth]{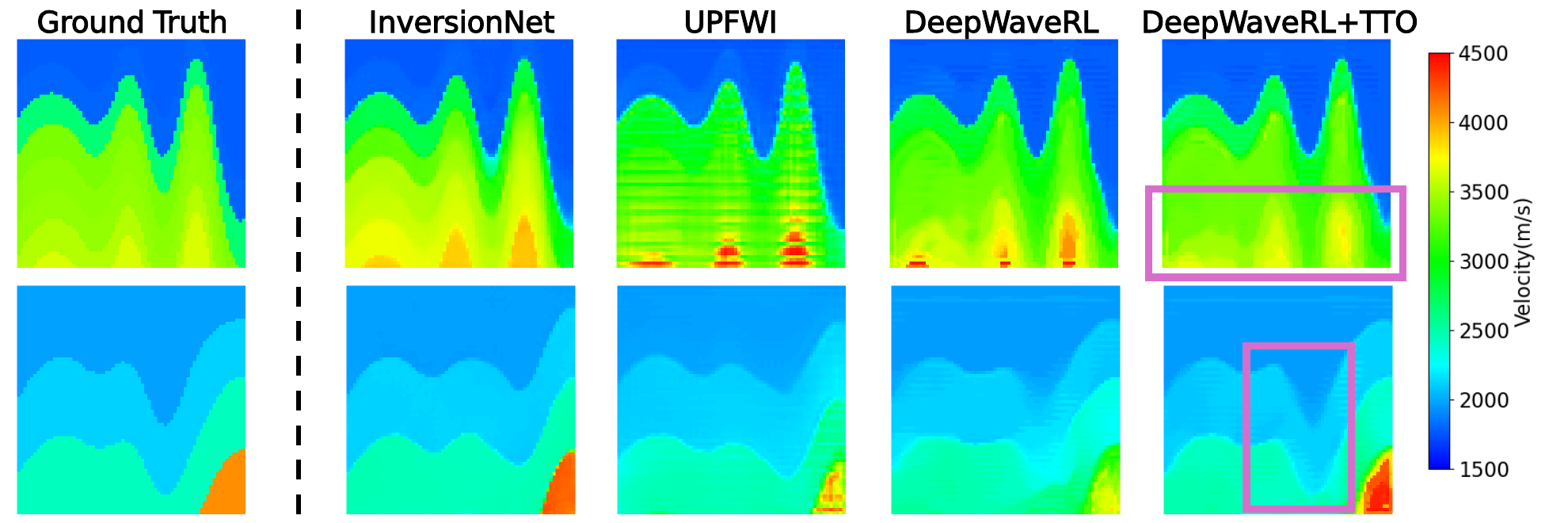}
    \caption{Illustration of ground truth and inversion results of different methods on CVA.}
    \label{fig:main}
    \vspace{-1.5em}
\end{figure*}

\subsection{Main Results}
\label{sec:main_results}

\textbf{Results on CVA: }Table~\ref{tab:main} shows the quantitative results of different methods on CVA. For the models that predict velocity as discrete values, we report the mean prediction (expected value) where we compute the expectation over the bin values using their predicted probabilities. 
Among all the models, InversionNet yields the best performance, which is expected as it leverages supervised learning and directly predicts continuous velocity values. In comparison, our DeepWaveRL with test-time optimization (DeepWaveRL+TTO) attains the second-best performance with a slight gap. Notably, even without TTO, DeepWaveRL achieves better performance compared to UPFWI, and TTO further provides a substantial improvement.  

Figure~\ref{fig:main} further illustrates examples of ground-truth velocity maps and predictions from different methods. Here, the results of DeepWaveRL models are all mean predictions. Consistent with our quantitative analysis, InversionNet produces sharp layer boundaries and smooth and uniform velocities within layers. The predictions of UPFWI contain artifacts in deep regions. In contrast, DeepWaveRL+TTO yields more details. As highlighted in the first row, the predictions of DeepWaveRL+TTO have more accurate velocities in deep regions, whereas InversionNet introduces additional layers and predicts inaccurate velocities. Another observation is that only DeepWaveRL+TTO precisely reconstructs the subsurface structure in the highlighted areas in the second row. Moreover, when comparing the results of DeepWaveRL and DeepWaveRL+TTO, we find that TTO helps eliminate artifacts, recover curved structures, and improve the accuracy of intra-layer velocities. More visualization results are shown in Appendix~\ref{fig:cva_sup}.

\begin{table*}[t]
\caption{Quantitative results evaluated on FFA and CFA. }
\label{tab:main_transfer}
\begin{center}
\setlength{\tabcolsep}{15pt}
\renewcommand{\arraystretch}{1.1}
\small
\begin{tabular}{c|c|c|c|c}
\Xhline{5\arrayrulewidth}
Dataset & Method & MAE$\downarrow$& RMSE$\downarrow$ & SSIM$\uparrow$ \\
\Xhline{5\arrayrulewidth}
\multirow{4}{*}{FFA} &  InversionNet &  \textbf{0.0098} & \textbf{0.0276} & \textbf{0.9880} \\ \cline{2-5}
& UPFWI & \uwave{0.0214} & 0.0484 & \uwave{0.9674}  \\ \cline{2-5}
& DeepWaveRL & 0.0301 & 0.0557 & 0.9062 \\ \cline{2-5}
& DeepWaveRL + TTO & 0.0268 & \uwave{0.0476} & 0.9146 \\  \hline

\multirow{4}{*}{CFA} &  InversionNet &  \textbf{0.0164} & \textbf{0.0480} & \textbf{0.9721} \\ \cline{2-5}
& UPFWI &  0.0280 & 0.0641 & \uwave{0.9492} \\ \cline{2-5}
& DeepWaveRL & 0.0362 & 0.0703 & 0.9111 \\ \cline{2-5}
& DeepWaveRL + TTO & \uwave{0.0278} & \uwave{0.0502} & 0.9323 \\ 
\Xhline{4\arrayrulewidth}
\end{tabular}
\end{center}
\vspace{-1em}
\end{table*}

\begin{figure*}
    \centering
    \includegraphics[width=1.\linewidth]{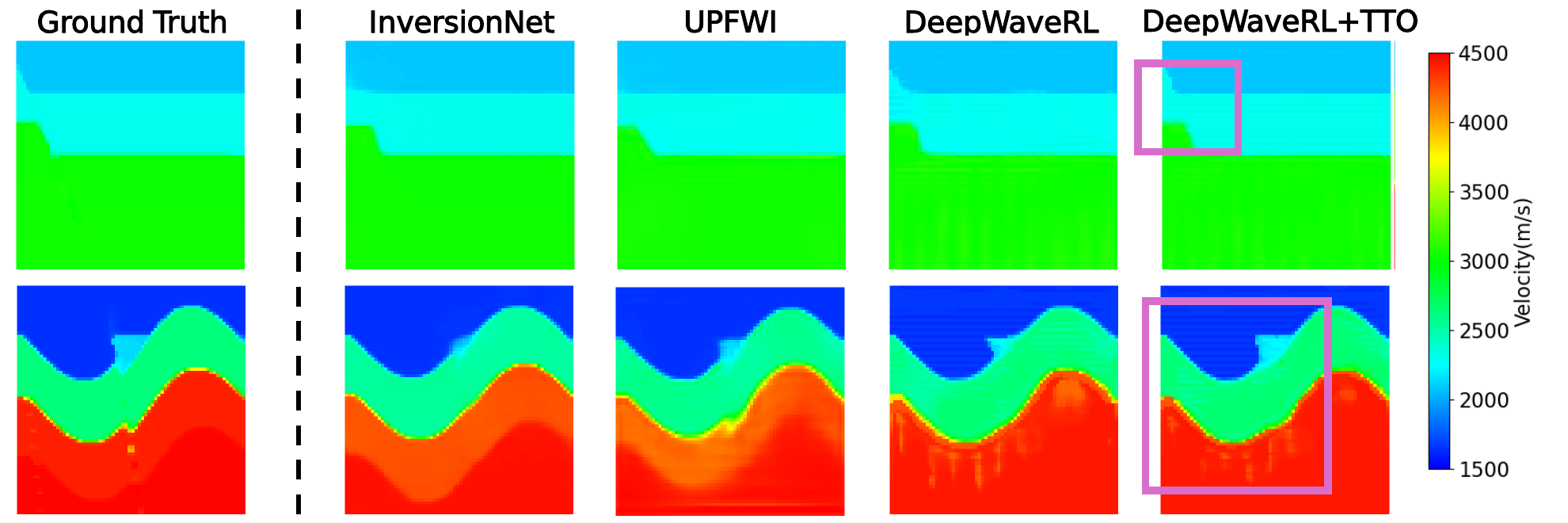}
    \caption{Illustration of ground truth and inversion results of different methods on FFA (top) and CFA (bottom).}
    \label{fig:main_transfer}
\end{figure*}

\textbf{Transfer Learning Results on FFA and CFA:} Table~\ref{tab:main_transfer} lists the quantitative results on FFA and CFA, and mean predictions are reported. For transfer learning, we adopt the last checkpoint of the DeepWaveRL+TTO on CVA. All the baseline models are trained from scratch. For FFA, our DeepWaveRL+TTO outperforms UPFWI in terms of RMSE. For CFA, it outperforms UPFWI in terms of MAE and RMSE, with small gaps in SSIM. This indicates that DeepWaveRL+TTO yields generally accurate predictions, but there may be small shifts or artifacts that are visually noticeable. 

The visualization results in Figure~\ref{fig:main_transfer} further support our hypothesis. Despite some stripe artifacts, our DeepWaveRL and DeepWaveRL+TTO generate more precise details in some shallow regions. As highlighted in the first row (FFA), DeepWaveRL models reconstruct the fault on the top-left corner, while it is barely visible in the predictions of InversionNet and UPFWI. Similarly, in the second row (CFA), our DeepWaveRL models precisely capture the triangle-shaped region. More visualization results are shown in Appendix~\ref{fig:fault_sup}.

We further report wall-clock time of training and inference and DeepWaveRL+TTO with different computation budgets in Appendix~\ref{sec:computation}. Transfer learning for baselines and UPFWI with TTO are also provided in Appendix~\ref{sec:trans_and_tto} for reference.

\subsection{Ablations}
\label{sec:ablation}
\begin{table*}[t]
\caption{Quantitative results evaluated on CVA, with different choices of action space.}
\label{tab:abl}
\begin{center}
\setlength{\tabcolsep}{15pt}
\renewcommand{\arraystretch}{1.1}
\small
\vspace{-1em}
\begin{tabular}{c|c|c|c}
\Xhline{4\arrayrulewidth}
Setting & MAE$\downarrow$& RMSE$\downarrow$ & SSIM$\uparrow$ \\
\Xhline{4\arrayrulewidth}
Continuous Action Space & 0.1443 & 0.2073 & 0.4359 \\ \hline
Discrete Action Space & \textbf{0.0527} & \textbf{0.1012} & \textbf{0.8601} \\
\Xhline{4\arrayrulewidth}
\end{tabular}
\end{center}
\vspace{-0.5em}
\end{table*}

\begin{figure*}[t]
    \centering
    \includegraphics[width=0.95\linewidth]{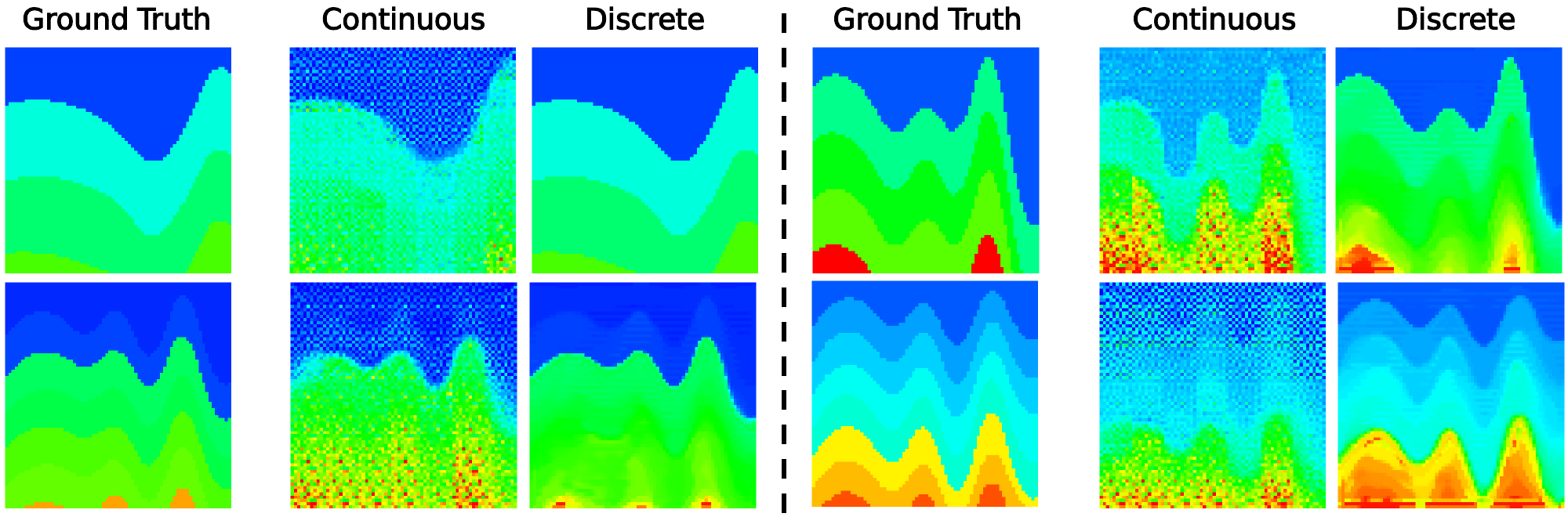}
    \caption{Illustration of ground truth and inversion results of DeepWaveRL with different choices of action space}
    \label{fig:ablation}
    \vspace{-1em}
\end{figure*}

\textbf{Continuous vs. Discrete Action Space: }We further analyze how the choice of action space affects model performance. The quantitative results are summarized in Table~\ref{tab:abl}. For simplicity, we denote DeepWaveRL with a discrete action space as DeepWaveRL-D, and with a continuous action space as DeepWaveRL-C. In terms of all three metrics, DeepWaveRL-D outperforms DeepWaveRL-C to a large extent. The gap is particularly large in SSIM, suggesting that DeepWaveRL-C fails to produce results consistent with human perceptual quality.

Figure~\ref{fig:ablation} provides qualitative comparisons between DeepWaveRL-C and DeepWaveRL-D, which further support our quantitative analysis. While DeepWaveRL-C can recover some structures in shallow regions, there are plenty of artifacts all over the predictions. In particular, the high-velocity areas are severely corrupted, making boundaries unrecognizable. Furthermore, these artifacts persist even when training is extended, indicating that the continuous action space poses significant optimization challenges. By contrast, discretization reduces the complexity of the action space and greatly stabilizes training, leading to more accurate and reliable results.

\textbf{Discretization Resolution:} To determine the optimal boundary for the exploration manifold, we evaluate DeepWaveRL across different discretization resolutions in terms of the number of bins $B$. We employ the last checkpoint of DeepWaveRL+TTO on CVA and append a $1\times 1$ convolutional layer as an adapter to output predictions of different bins. The quantitative results listed in Table~\ref{tab:bins} show a trade-off between physical resolution and optimization difficulties. The visualization results are depicted in Appendix~\ref{fig:bins}.

\textbf{Logarithm Transformation:} To evaluate the effect of the sign-preserving logarithm transformation, we train our DeepWaveRL without transformation on CVA. The resulting MAE, RMSE, and SSIM are 0.0785, 0.1383, and 0.8195, respectively. Compared to the results of DeepWaveRL with transformation in Table~\ref{tab:main}, the performance degrades in terms of all three metrics. This is consistent with the visualization results in Appendix~\ref{fig:wo_log}, where some details are missing in deep regions.

\textbf{Group Size Scaling:} To empirically analyze how group size $G$ affects the gradient signal-to-noise ratio, we train DeepWaveRL with various group sizes on CVA from scratch. The quantitative results are listed in Table~\ref{tab:group}. As expected theoretically, the large group size ($G=256$) is required to suppress the variance inherent to high-dimensional spatial RL, ensuring the advantage estimator reliably differentiates between spatially coherent and incoherent actions. The visualization results are shown in Appendix~\ref{fig:group}.

\textbf{Prior as Regularization:} To analyze the importance of the structural prior to stabilize training, we directly train DeepWaveRL on the test set of CVA, starting from random initialization. Without the prior, DeepWaveRL yields significantly degraded performance. The resulting MAE, RMSE and SSIM are 0.2120, 0.3004, and 0.5832, respectively. Additionally, we transfer the checkpoint on CVA to the test set of FFA and CFA without fine-tuning on the training set. The quantitative results in Table~\ref{tab:tto_direct} indicate that fine-tuning helps refine the prior. The visualization results are shown in Appendix~\ref{fig:tto_direct}.

\begin{table}[t]
\centering
\setlength{\tabcolsep}{15pt}
\renewcommand{\arraystretch}{1.0}
\small
\caption{Comparison of DeepWaveRL+TTO with and without fine-tuning on the training set of FFA and CFA.}
\label{tab:tto_direct}
\begin{tabular}{c|c|c|c|c}
\Xhline{4\arrayrulewidth}
Dataset & Fine-tuning & MAE $\downarrow$ & RMSE $\downarrow$ & SSIM $\uparrow$ 
\\ \Xhline{4\arrayrulewidth}
\multirow{2}{*}{FFA} & \checkmark  &  \textbf{0.0268} & \textbf{0.0476} & 0.9146 \\ \cline{2-5}
&  & 0.0287 &  0.0520& \textbf{0.9347} \\ \hline
\multirow{2}{*}{CFA} & \checkmark & \textbf{0.0278} & \textbf{0.0502} & \textbf{0.9323} \\ \cline{2-5}
&  & 0.0358 & 0.0647 & 0.9153 \\
\Xhline{4\arrayrulewidth}
\end{tabular}
\vspace{-1em}
\end{table}

\begin{table}[t]
    \centering
    \begin{minipage}{0.48\textwidth}
        \centering
        \setlength{\tabcolsep}{8pt}
        \renewcommand{\arraystretch}{1.0}
        \small
        \caption{Comparison of DeepWaveRL+TTO with different numbers of bins on CVA.}
        \label{tab:bins}
        \vspace{0.3em}
        \begin{tabular}{c|c|c|c}
        \Xhline{4\arrayrulewidth}
        \# Bins & MAE $\downarrow$ & RMSE $\downarrow$ & SSIM $\uparrow$ 
        \\ \Xhline{4\arrayrulewidth}
        50 & 0.0546 & 0.1027 & 0.8493 \\ \hline
        100 & 0.0527 & 0.1012 & 0.8601 \\ \hline
        200 & \textbf{0.0519} & \textbf{0.1010} & \textbf{0.8617} \\ \hline
        500 & 0.0521 & 0.1010 & 0.8609 \\ 
        \Xhline{4\arrayrulewidth}
        \end{tabular}
    \end{minipage}%
    \hfill
    \begin{minipage}{0.48\textwidth}
        \centering
        \setlength{\tabcolsep}{6pt}
        \renewcommand{\arraystretch}{1.0}
        \small
        \caption{Comparison of DeepWaveRL with different group sizes on CVA.}
        \label{tab:group}
        \vspace{0.3em}
        \begin{tabular}{c|c|c|c}
        \Xhline{4\arrayrulewidth}
        Group Size & MAE $\downarrow$ & RMSE $\downarrow$ & SSIM $\uparrow$ 
        \\ \Xhline{4\arrayrulewidth}
        32 & 0.0948  & 0.1535 & 0.7878 \\ \hline
        64 & 0.0796 & 0.1324 & 0.8214 \\ \hline
        128 & 0.0885 & 0.1680 & 0.8151 \\ \hline
        256 & \textbf{0.0717} & \textbf{0.1300} & \textbf{0.8303} \\ 
        \Xhline{4\arrayrulewidth}
        \end{tabular}
    \end{minipage}
    \vspace{-1em}
\end{table}



\section{Discussion}
During qualitative analysis, we observe that mode collapse occurs in the predictions of DeepWaveRL (\textasciitilde 1.85\% of test cases), as illustrated in Appendix~\ref{fig:fail}. This ratio decreases to 1.73\% after TTO. There is no consistent pattern in velocity maps or seismic data that guarantees the occurrence of mode collapse. We plan to dive deeper to solve this issue in our future work. Another limitation is that DeepWaveRL still cannot perfectly recover the structures in deeper regions due to the inherent attenuation of signals in these regions. Our sign-preserving logarithm transformation is one of the solutions, but more advanced algorithms are still needed. Furthermore, our DeepWaveRL framework enables another potential research direction, which is to incorporate conventional regularization terms such as total variation into the reward design.


\section{Related Work}
\textbf{Deep Learning for FWI:~}Deep learning approaches to FWI span data-driven, physics-informed, and hybrid paradigms~\citep{Physics-2023-Lin, Deep-2021-Adler, Deep-2021-Yu}. Fully supervised methods~\citep{araya2018deep, wu2019inversionnet, zhang2019velocitygan, li2020c} learn direct mappings from seismic data to velocity models using paired data, which are costly to acquire and often lead to poor generalization under domain shifts. To reduce reliance on labels and improve robustness, self-supervised strategies have emerged. Feng et al.~\citep{feng2022intriguing, feng2024auto} decouple the seismic encoder and velocity decoder by leveraging latent space correlations, enabling separate training. SiameseFWI~\citep{SiameseFWI-2024-Saad} explores self-supervision with a Siamese network that better aligns simulated and observed data. Semi-supervised learning has also been explored. \cite{Learning-2023-Sun} proposes a CycleGAN-based framework to reconstruct missing low-frequency components in field data. Other methods generate pseudo-labels from unlabeled or auxiliary data~\citep{Renan-2022-Physics, Cai-2022-Semi}, bridging the gap between labeled and unlabeled domains. Unsupervised methods such as UPFWI~\citep{jin2021unsupervised} and Jia et al.~\citep{Seismic-2025-Jia} go further by eliminating labels entirely. These approaches minimize waveform mismatches under physical constraints, using differentiable forward modeling to optimize predicted velocity maps. However, their reliance on computationally intensive and differentiable solvers limits scalability to high-resolution, elastic, or 3D FWI, and precludes use with black-box simulators. Diffusion models offer an alternative by learning generative priors that guide the inverse process via plug-and-play (PnP) denoising~\citep{song2022solving, chung2023diffusion, zhang2025improving}. \cite{Prior-2023-Wang} successfully applies this strategy to FWI. While these models avoid paired supervision, they still require large velocity datasets for training and incur high inference costs due to repeated forward simulations during sampling.

\textbf{Reinforcement learning and group-based policy optimization:}
Policy-gradient RL methods (e.g., PPO~\citep{schulman2017proximal}) provide a principled way to optimize stochastic policies via likelihood-ratio estimators, and have been widely used in sequential generation and control. Recent advances in group-/sequence-level policy optimization demonstrate that performing importance-weighting and clipping at the unit-of-reward level (group or sequence) can reduce variance and stabilize training for structured outputs. In particular, DeepSeekMath~\citep{deepseek-math} replaces value-function estimation with group-relative normalization, yielding more efficient and stable updates. They show that this framework can be scaled to mathematical reasoning tasks with strong generalization. DAPO~\citep{yu2025dapo} extends the paradigm by decoupling clipping ranges and introducing dynamic sampling, further improving stability under diverse reward distributions. GSPO~\citep{zheng2025group} generalizes these ideas to full sequence-level optimization with the importance ratio based on sequence likelihood, aligning long-horizon objectives with token-level policies. Inspired by this progression, DeepWaveRL adapts the same philosophy to the geophysics domain by treating an entire velocity map as a structured action, making map-level optimization a natural extension of group-based RL methods for physics-driven inverse problems.

\section{Conclusion}
In this study, we introduce DeepWaveRL, a reinforcement learning framework for self-supervised full waveform inversion under the strict scalar-reward regime. By integrating a map-level advantage estimator, discretized spatial actions, a sign-preserving logarithmic reward transformation, and prior regularization via transfer learning, DeepWaveRL overcomes the massive combinatorial action space and spatial credit assignment bottlenecks inherent to the high-dimensional velocity domain. Our empirical analysis demonstrates that these strategies effectively bound the exploration manifold and suppress policy gradient variance, enabling DeepWaveRL to achieve competitive structural fidelity compared to both supervised and self-supervised baselines. This approach provides a flexible paradigm for FWI, offering new possibilities in real-world settings.

\bibliographystyle{abbrvnat}
\bibliography{references}


\appendix

\section{Comparison with Previous FWI Methods from a Gradient Perspective}
\label{sec:gradient_comparison}

To elucidate the relationship among supervised, exact-gradient self-supervised, and our scalar-reward reinforcement learning based self-supervised approaches, we provide a unified analysis from the perspective of gradient construction and propagation.

\paragraph{Supervised Learning:}
In the supervised paradigm, a neural network $g_\theta$ maps seismic measurements $p$ to a predicted velocity map $\hat{V} = g_\theta(p)$. Training relies on paired ground-truth velocity maps $v$, with a loss function of the form
\begin{equation}
    \mathcal{L}_{\mathrm{sup}}(\theta) 
    \;=\; \mathbb{E}_{(p,V)\sim \mathcal{D}}
           \,\|\, V - g_\theta(p) \,\| \,,
\end{equation}
where $\mathcal{D}$ denotes the joint distribution of seismic data and corresponding velocity maps, and $\|\cdot\|$ denotes a generic norm (e.g., $\ell_1$, $\ell_2$, or mixed/perceptual norms). 
The gradient is obtained via direct backpropagation:
\begin{equation}
    \nabla_\theta {\mathcal{L}_{\text{sup}}}
    = \mathbb{E}_{(p,V)\sim \mathcal{D}}
    \frac{\partial {\mathcal{L}_{\text{sup}}}}{\partial \hat{V}} \cdot 
    \frac{\partial \hat{V}}{\partial \theta}
    = \mathbb{E}_{(p,V)\sim \mathcal{D}}
    \textcolor{blue}{\underbrace{\textcolor{black}{\frac{\partial \textcolor{blue}{\mathcal{L}_{\text{sup}}}}{\partial g_\theta(p)}}}_{\text{Ground truth involved}}} \cdot 
    \frac{\partial g_\theta(p)}{\partial \theta},
\end{equation}

where the learning signal is explicitly anchored to the availability of ground-truth velocity maps. While this yields stable optimization and strong supervision, it is impractical in real-world scenarios due to the scarcity of paired data.

\paragraph{Exact-Gradient Self-Supervised Learning:}
The UPFWI framework~\citep{jin2021unsupervised} removes the dependence on ground truth by incorporating a differentiable forward operator $f$ that simulates seismic data from predicted velocity maps:
\begin{equation}
    \hat{V} = g_\theta(p), \quad \tilde{p} = f(\hat{V}).
\end{equation}
The reconstruction objective is defined as
\begin{equation}
    \mathcal{L}_{\mathrm{diff}}(\theta) = \mathbb{E}_{p\sim \mathcal{P}}\|p - \tilde{p}\| = \mathbb{E}_{p\sim \mathcal{P}}\|p - f(g_\theta(p))\|,
\end{equation}
with gradients computed via the chain rule:
\begin{equation}
    \nabla_\theta \mathcal{L}_{\mathrm{diff}}
    = \mathbb{E}_{p\sim \mathcal{P}}\frac{\partial \mathcal{L}_{\mathrm{diff}}}{\partial \tilde{p}}
      \cdot \frac{\partial \tilde{p}}{\partial \hat{V}}
      \cdot \frac{\partial \hat{V}}{\partial \theta}
    = \mathbb{E}_{p\sim \mathcal{P}}\frac{\partial \mathcal{L}_{\mathrm{diff}}}{\partial f(g_\theta(p))}
      \cdot \textcolor{blue}{\underbrace{\frac{\partial f(g_\theta(p))}{\partial g_\theta(p)}}_{\text{Differentiable}}}
      \cdot \frac{\partial g_\theta(p)}{\partial \theta}.
\end{equation}
This formulation enables end-to-end training using only seismic data, yet it couples the optimization tightly to the availability of a differentiable solver to compute $\partial f / \partial \hat{V}$.

\paragraph{Scalar-Reward Self-Supervised Learning via Reinforcement Learning:}
Our proposed DeepWaveRL frames FWI as a policy optimization problem. With the definition in the above sections, we can derive the gradient of our objective as follows (clipping is omitted for brevity):
\begin{align}
    \nabla_\theta \mathcal{J}(\theta)
    &= \nabla_\theta \mathbb{E}_{{p\sim \mathcal{P}, \{\hat{V}\}_{i=1}^G\sim \pi_{\theta_\text{old}}(\cdot \mid p)}} \frac{1}{G}\sum_{i=1}^{G} \Big\{m_{i}(\theta) \hat{A}_i \Big\} \\
    &= \mathbb{E}_{{p\sim \mathcal{P}, \{\hat{V}\}_{i=1}^G\sim \pi_{\theta_\text{old}}(\cdot \mid p)}}\!\Bigg[\,
           \frac{1}{G}\sum_{i=1}^G m_i(\theta)\,\hat{A}_i \cdot \; \nabla_\theta \log m_i(\theta) \,\Bigg] \\
    &= \mathbb{E}_{{p\sim \mathcal{P}, \{\hat{V}\}_{i=1}^G\sim \pi_{\theta_\text{old}}(\cdot \mid p)}}\! \\
    &\qquad\qquad\qquad \underset{\textcolor{blue}{\text{Forward model only involved as a multiplier}}}{\Bigg[\,
           \frac{1}{G}\sum_{i=1}^G \left(\frac{\pi_\theta(\hat{V}_i\mid p)}{\pi_{\theta_{\mathrm{old}}}(\hat{V}_i\mid p)}\right)^{\!\frac{1}{H\cdot W}}
           \textcolor{blue}{\underbrace{\hat{A}_i}} \cdot\frac{1}{H\cdot W}\sum_{h,w} \nabla_\theta \log \pi_\theta(\hat{v}_{i,h,w}\mid p) \,\Bigg]}.
\end{align}
Unlike supervised and differentiable self-supervised learning, where the loss must be differentiable with respect to the high-dimensional spatial output $\hat{V}$ and thus couples optimization tightly to the properties of $f$, reinforcement learning completely severs the dependency and transforms the reconstruction error into a scalar reward signal that directly re-weight policy gradients.
%

Thus, the three paradigms can be interpreted within a common gradient-based framework:
\textbf{Supervised:} Gradients are directly computed from discrepancies with labeled velocity maps, yielding stable optimization but requiring costly ground truth.
\textbf{Exact-gradient self-supervised:} Gradients originate from seismic reconstruction error and backpropagate through a differentiable forward operator, enabling physics-guided learning but restricted by differentiability.
\textbf{Scalar-reward RL-based self-supervised:} Gradients arise from log-likelihood weighting in policy space, with seismic misfit entering only as a reward. This bypasses the need for exact physics gradients entirely, and enables stochastic exploration in complex inversion landscapes.

\section{Training Details}
\label{sec:tech}
We normalize the input seismic data to the range $[-1,1]$ and apply the logarithm transformation with $k=3$ and $c=0$ on seismic data. For optimization, we employ the AdamW optimizer with momentum parameters $\beta_1=0.9$, $\beta_2=0.999$, and a weight decay of $1\times10^{-4}$ to update all parameters of the network. The details of hyperparameters and training settings are provided in Table~\ref{tab:train}. The $\epsilon_\text{low}$ and $\epsilon_\text{high}$ are 0.2 and 0.27, respectively. For the network architecture, we adopt a four-layer encoder-decoder Vision Transformer (ViT), and we append four convolutional blocks with upsampling layers ($5\times$ and $2\times$), batch normalization, and leaky ReLU as activation functions to map the output of the decoder to $70\times 70$ velocity map with $100$ bins. We implement our models in Pytorch and train them on 16 NVIDIA H100 GPUs. 

\begin{table}[h]
\centering
\setlength{\tabcolsep}{15pt}
\renewcommand{\arraystretch}{1.1}
\caption{Training details}
\label{tab:train}
\small
\begin{tabular}{c|c|c|c|c}
\Xhline{4\arrayrulewidth}
                        & \multicolumn{2}{c|}{CVA} & \multicolumn{2}{c}{FFA \& CFA} \\
                    \Xhline{4\arrayrulewidth}
Test-time Optimization &            & $\checkmark$         &                & $\checkmark$             \\ \hline
Training Steps          & 44,880     & 1,440      & 7,360          & 1,600         \\ \hline
Initial Learning Rate   & 8e-4       & 1.6e-4     & 6.4e-4         & 6.4e-4        \\ \hline
Learning Rate Decay     & /          & 1,360      & /              & /             \\ \hline
Batch Size              & 128        & 2048       & 2048           & 2048          \\ \hline
Group Size              & 256        & 16         & 32             & 32 \\           
\Xhline{4\arrayrulewidth}
\end{tabular}
\end{table}

\section{Wall-Clock Time and Performance with Various Budgets}
\label{sec:computation}
The wall-clock training times are summarized in Table~\ref{tab:wall_train}, and the single-pass wall-clock inference times are listed in Table~\ref{tab:wall_test}. The TTO process takes around 1 hour and 15 minutes for CVA and 2 hours 40 minutes for FFA/CFA on 16xH100. Considering the number of test samples (6000), each sample in CVA takes approximately 0.75 seconds, and the one in FFA/CFA takes 1.6 seconds. For TTO on CVA, there are 11,520 forward simulations per sample (720 epochs). For FFA/CFA, there are 25,600 (800 epochs).

\begin{figure}[h!]
    \centering
    \includegraphics[width=0.98\linewidth]{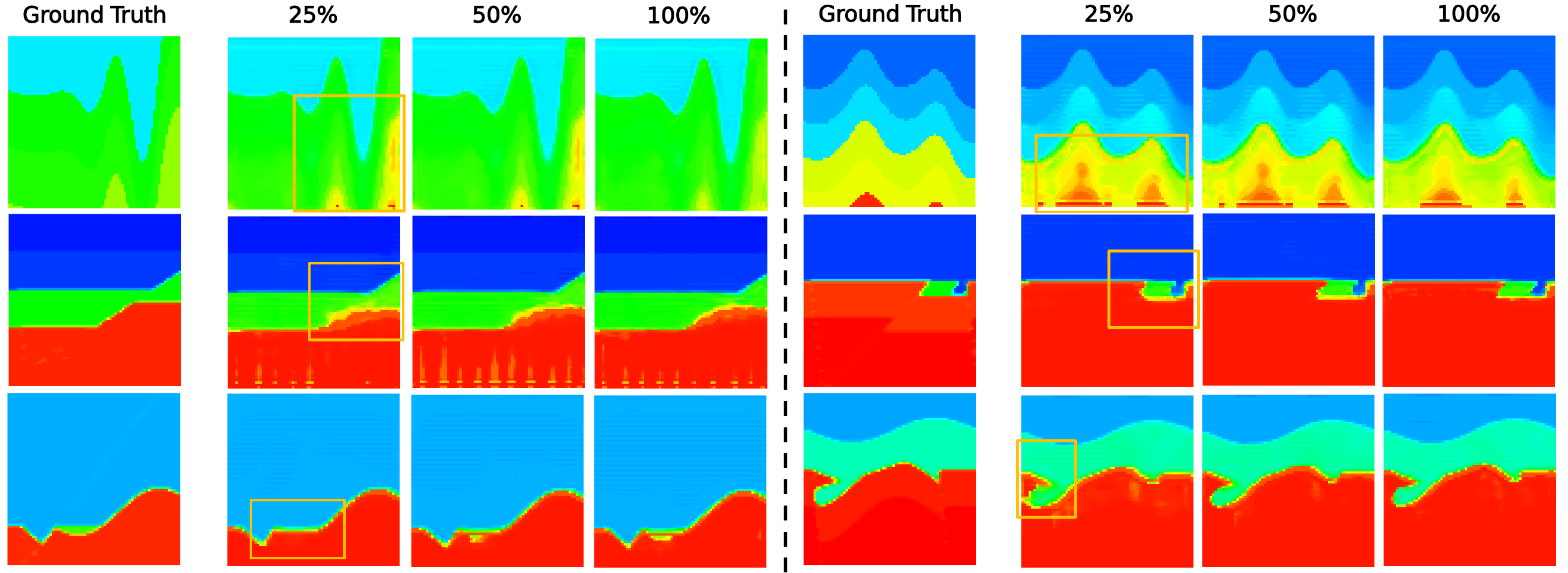}
    \caption{Comparison of DeepWaveRL+TTO with different computation budgets on CVA (top row), FFA (middle row) and CFA(bottom row). }
    \label{fig:budget}
\end{figure}

We further evaluate the model performance at intermediate TTO milestones to demonstrate the performance of DeepWaveRL+TTO with different computation budgets. The quantitative results are shown in Table~\ref{tab:budget}, the visualized results are shown in Figure~\ref{fig:budget}.

\begin{table}[h!]
\centering
\setlength{\tabcolsep}{15pt}
\renewcommand{\arraystretch}{1.1}
\caption{Wall-clock training time.}
\label{tab:wall_train}
\small
\begin{tabular}{c|c|c}
\Xhline{4\arrayrulewidth}
Method (Hardware) & CVA & FFA/CFA \\ \Xhline{4\arrayrulewidth}
InversionNet (1xH100) & 51 mins & 1 hr 30 mins\\ \hline
UPFWI (4xH100) & 14 hrs 45 mins& 27 hrs\\ \hline
DeepWaveRL (16xH100) & 27 hrs& 11 hrs \\      
\Xhline{4\arrayrulewidth}
\end{tabular}
\end{table}

\begin{table}[h!]
\centering
\setlength{\tabcolsep}{15pt}
\renewcommand{\arraystretch}{1.1}
\caption{Wall-clock inference time (single pass inference).}
\label{tab:wall_test}
\small
\begin{tabular}{c|c|c}
\Xhline{4\arrayrulewidth}
Method & Mean & Std \\ \Xhline{4\arrayrulewidth}
DeepWaveRL &  0.3773 & 0.8622 \\ \hline
UPFWI &  0.3975 & 0.9789 \\ \hline
InversionNet & 0.3381 & 0.9418 \\
\Xhline{4\arrayrulewidth}
\end{tabular}
\end{table}

\begin{table}[h!]
\centering
\setlength{\tabcolsep}{15pt}
\renewcommand{\arraystretch}{1.1}
\caption{Comparison of DeepWaveRL+TTO with different computation budgets.}
\label{tab:budget}
\small
\begin{tabular}{c|c|c|c|c}
\Xhline{4\arrayrulewidth}
Dataset & Epoch & MAE $\downarrow$ & RMSE $\downarrow$ & SSIM $\uparrow$ 
\\ \Xhline{4\arrayrulewidth}
\multirow{3}{*}{CVA} 
 & 720 & \textbf{0.0527} & \textbf{0.1012} & \textbf{0.8601} \\\cline{2-5}
 & 360 & 0.0569 & 0.1075 & 0.8524 \\ \cline{2-5}
 & 180 & 0.0676 & 0.1172 & 0.8455 \\ \hline
\multirow{3}{*}{FFA} 
& 800 & \textbf{0.0268} & 0.0476 & \textbf{0.9146} \\ \cline{2-5}
& 400 & 0.0270 & \textbf{0.0471} & 0.9143 \\ \cline{2-5}
 & 200 & 0.0277 & 0.0487 & 0.9134 \\ \hline
\multirow{3}{*}{CFA} 
& 800 & \textbf{0.0278} & \textbf{0.0502} & \textbf{0.9323} \\ \cline{2-5}
& 400 & 0.0291 & 0.0527 & 0.9295 \\ \cline{2-5}
 & 200 & 0.0307 & 0.0564 & 0.9255 \\ \hline
\Xhline{4\arrayrulewidth}
\end{tabular}
\end{table}

\section{Derivative-Free Optimization Baseline}
\label{sec:zero}

We implement simultaneous perturbation stochastic approximation (SPSA) as the derivative-free optimization  baseline on a subset of 100 samples from CVA test set.

First, we use a constant velocity of 2500 m/s as the initial model and optimize it with a learning rate of 50,000 and a perturbation magnitude of 100. The optimization is performed for 12,000 iterations per sample (24,000 forward simulations), which is approximately twice simulations of DeepWaveRL+TTO. We observe that SPSA struggled to converge to a physically reasonable velocity map. The resulting MAE, RMSE and SSIM are 0.5020, 0.6444, and 0.1457, respectively.

\begin{figure}[h]
    \centering
    \includegraphics[width=0.8\linewidth]{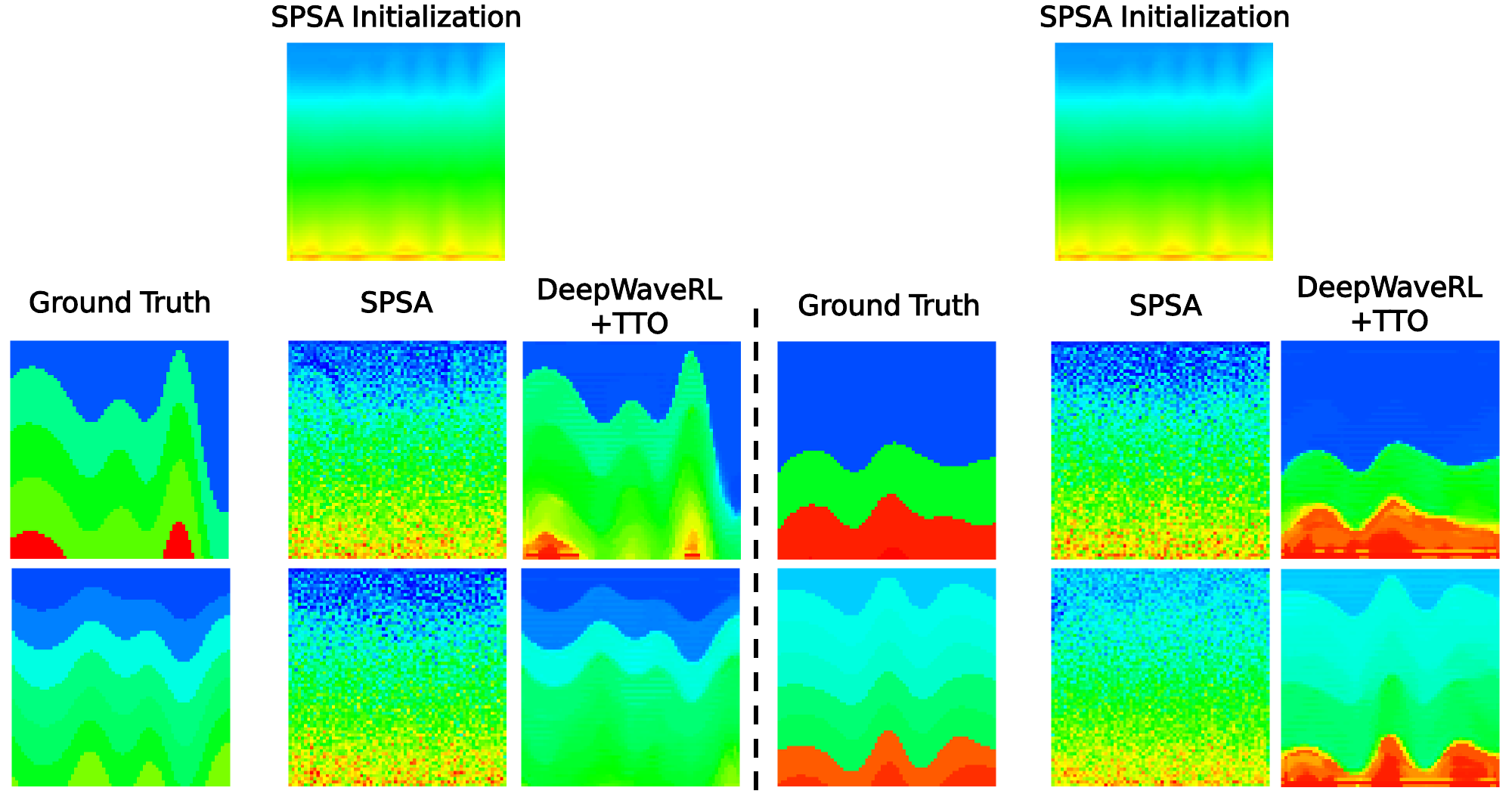}
    \caption{Illustration of ground truth and inversion results of SPSA and DeepWaveRL+TTO on CVA}
    \label{fig:spsa}
\end{figure}

We then use the mean prediction of DeepWaveRL on CVA training set as the warm-started initial model (as shown in Figure~\ref{fig:spsa}). The performance of SPSA improves (MAE:0.3590, RMSE:0.4890, SSIM:0.2226) compared to the one with constant initialization. These results strongly highlight the fundamental difficulty of optimizing a high-dimensional structured space using generic zeroth-order methods. 

\textbf{Comparison of DeepWaveRL and Perturbation-Based Optimizers:} Under the scalar-reward regime, the classical alternatives to our method are derivative-free optimization algorithms such as Simultaneous Perturbation Stochastic Approximation (SPSA) or Evolutionary Strategies (e.g., CMA-ES). While these paradigms are well-established, they face severe scalability bottlenecks when applied to the high-dimensional, structured output space of FWI as they typically perturb the space directly.

In contrast, our DeepWaveRL fundamentally changes the optimization target. Rather than directly perturbing independent velocity pixels, DeepWaveRL optimizes the weights of a deep policy network. The Vision Transformer and convolutional layers impose strong spatial inductive biases, inherently restricting the exploration manifold to physically coherent, spatially smooth structures. Therefore, while both DeepWaveRL and traditional methods reply on the scalar data misfit without accessing gradients from the forward operator, RL provides a vastly superior mechanism for navigating high-dimensional structured physics problems by leveraging the representational power and architectural priors of deep neural networks.

\section{Baselines with Transfer Learning and TTO}
\label{sec:trans_and_tto}

For baselines with transfer learning, we initialize both InversionNet and UPFWI with their CVA pre-trained weights and fine-tune them on FFA and CFA. The results are shown in Table~\ref{tab:transfer_base}. It is observed that transferring the learned priors yields improved performance for both baselines. However, our core conclusions still hold that DeepWaveRL+TTO remains high competitive with UPFWI on FFA and outperforms UPFWI on CFA in terms of both MAE and RMSE.

\begin{table}[h]
\centering
\setlength{\tabcolsep}{15pt}
\renewcommand{\arraystretch}{1.1}
\caption{Comparison of different models transferred from CVA.}
\label{tab:transfer_base}
\small
\begin{tabular}{c|c|c|c|c}
\Xhline{4\arrayrulewidth}
Dataset & Method & MAE $\downarrow$ & RMSE $\downarrow$ & SSIM $\uparrow$ 
\\ \Xhline{4\arrayrulewidth}
\multirow{3}{*}{FFA} & InversionNet &  \textbf{0.0087} & \textbf{0.0252} & \textbf{0.9893} \\ \cline{2-5}
& UPFWI & \uwave{0.0200} &  \uwave{0.0429} & \uwave{0.9738} \\ \cline{2-5}
& DeepWaveRL+TTO & 0.0268 & 0.0476 & 0.9146 \\ \hline
\multirow{3}{*}{CFA} & InversionNet & \textbf{0.0155} & \textbf{0.0467} & \textbf{0.9731} \\ \cline{2-5}
& UPFWI & 0.0300 & 0.0895  & 0.9514 \\ \cline{2-5}
& DeepWaveRL+TTO & \uwave{0.0278} & \uwave{0.0502} & \uwave{0.9323} \\ 
\Xhline{4\arrayrulewidth}
\end{tabular}
\end{table}

We further provide the quantitative results of UPFWI with TTO in Table~\ref{tab:upfwi_tto}. While we provide these results for illustration, we respectfully emphasize that comparing DeepWaveRL+TTO against UPFWI+TTO bridges two fundamentally different problem settings. Optimizing UPFWI per-sample at test time explicitly requires backpropagating exact, dense gradients through a differentiable forward operator (effectively reverting to classical, gradient-based iterative FWI). In contrast, the foundational premise of DeepWaveRL is to operate in the single scalar reward environment without accessing gradients from the forward operator.

\begin{table}[t]
\centering
\setlength{\tabcolsep}{15pt}
\renewcommand{\arraystretch}{1.1}
\caption{Comparison of UPFWI with and without TTO (and with transfer learning from CVA) and DeepWaveRL+TTO.}
\label{tab:upfwi_tto}
\small
\begin{tabular}{c|c|c|c|c}
\Xhline{4\arrayrulewidth}
Dataset & Method & MAE $\downarrow$ & RMSE $\downarrow$ & SSIM $\uparrow$ 
\\ \Xhline{4\arrayrulewidth}
\multirow{3}{*}{CVA} & UPFWI & 0.0739 & 0.1369 & 0.7420 \\ \cline{2-5}
& UPFWI+TTO & 0.0666 & 0.1243 & 0.7634 \\ \cline{2-5}
& DeepWaveRL+TTO & \textbf{0.0527} & \textbf{0.1012} & \textbf{0.8601} \\ \hline

\multirow{4}{*}{FFA} & UPFWI & 0.0214 & 0.0484 & 0.9676 \\ \cline{2-5}
& UPFWI+TTO & \textbf{0.0176} & \textbf{0.0364} & \textbf{0.9778} \\ \cline{2-5}
& UPFWI+TTO (CVA init.) & \textbf{0.0167} & \textbf{0.0331} & \textbf{0.0.9800} \\ \cline{2-5}
& DeepWaveRL+TTO & 0.0268 & 0.0476 & 0.9146 \\ \hline

\multirow{4}{*}{CFA} & UPFWI & 0.0280 & 0.0641 & 0.9492 \\ \cline{2-5}
& UPFWI+TTO & \textbf{0.0231} & \textbf{0.0494} & \textbf{0.9639} \\ \cline{2-5}
& UPFWI+TTO (CVA init.) & 0.0278 & 0.0859 & 0.9563 \\ \cline{2-5}
& DeepWaveRL+TTO & 0.0278 & 0.0502 & 0.9323 \\ \hline
\Xhline{4\arrayrulewidth}
\end{tabular}
\end{table}

\section{Robustness Test}

\begin{figure}[h]
    \centering
    \includegraphics[width=0.85\linewidth]{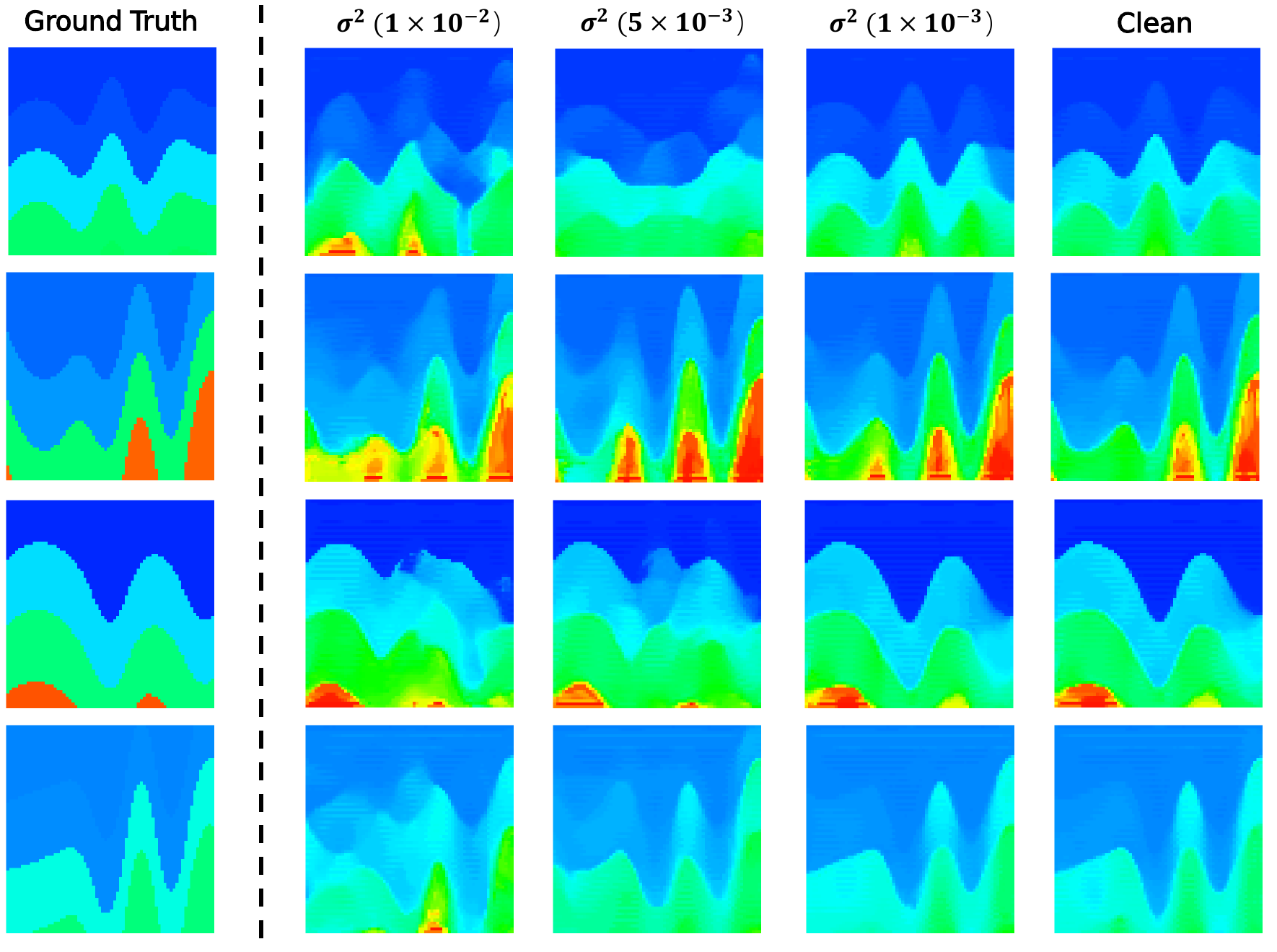}
    \caption{Comparison of DeepWaveRL+TTO tested on seismic data with different noise levels. }
    \label{fig:noise}
\end{figure}

We additionally validate the robustness of DeepWaveRL on CVA by adding different levels of gaussian noise to seismic data. The quantitative results listed in Table~\ref{tab:noise} indicate that our DeepWaveRL is robust to a certain level of noise. The visualization results are shown in Figure~\ref{fig:noise}

\begin{table}[h]
\centering
\setlength{\tabcolsep}{15pt}
\renewcommand{\arraystretch}{1.1}
\caption{Performance of DeepWaveRL+TTO tested on seismic data with different noise levels.}
\label{tab:noise}
\small
\begin{tabular}{c|c|c|c|c}
\Xhline{4\arrayrulewidth}
$\sigma^2 (\times 10^{-3})$ & PSNR & MAE $\downarrow$ & RMSE $\downarrow$ & SSIM $\uparrow$  \\
\Xhline{4\arrayrulewidth}
0 & / & 0.0527 & 0.1012 & 0.8601 \\ \hline
1 & 30.00 & 0.0593 & 0.1104 & 0.8446 \\ \hline
5 & 23.01 & 0.0907 & 0.1504 & 0.7931 \\ \hline 
10 & 20.00 & 0.1358 & 0.2090 & 0.7418 \\
\Xhline{4\arrayrulewidth}
\end{tabular}
\end{table}

\clearpage

\section{Visualizations}
\subsection{More Visualization Results on CVA}
\label{fig:cva_sup}
\begin{figure}[h]
    \centering
    \includegraphics[width=0.78\linewidth]{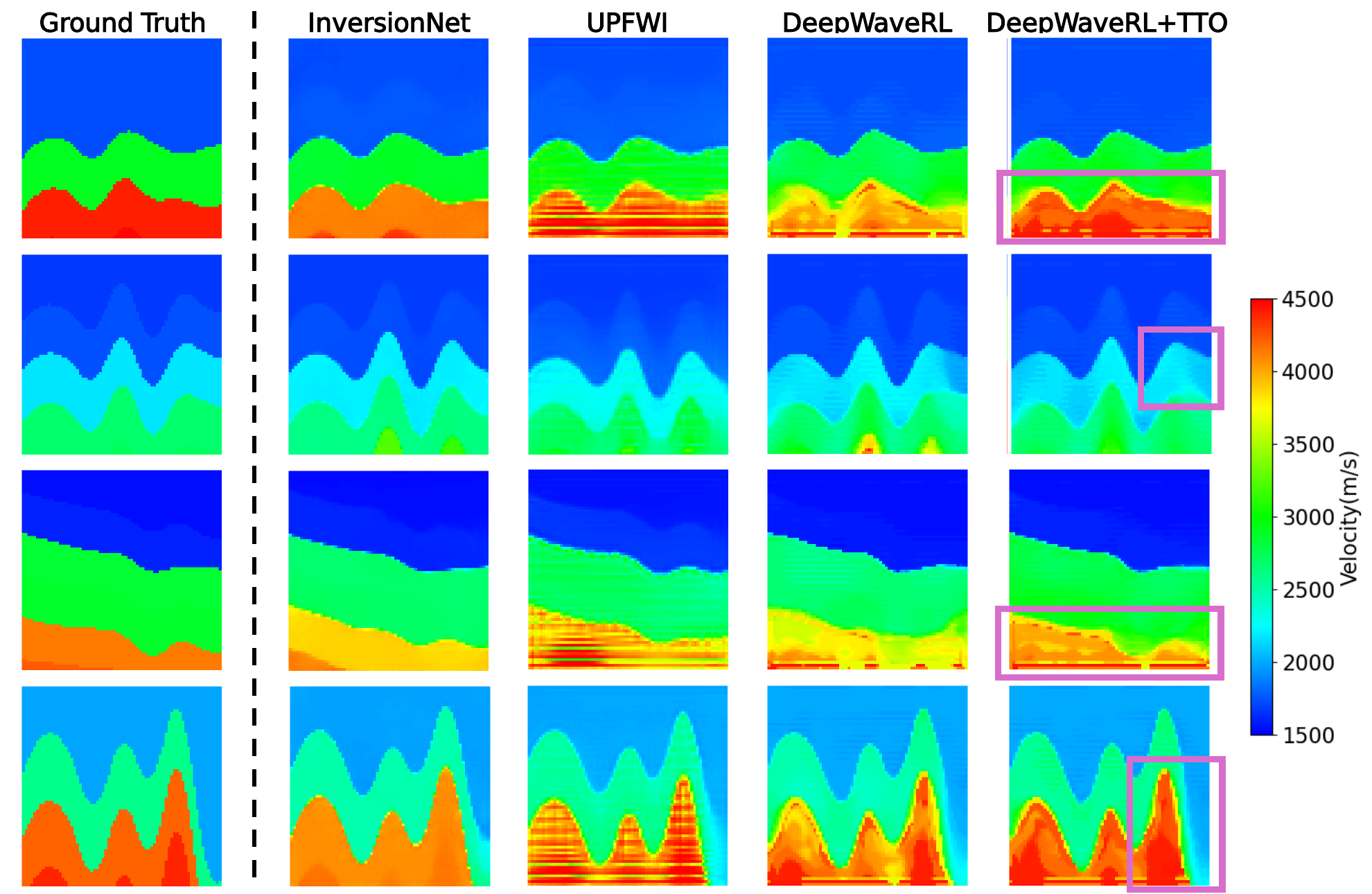}
    \caption{Illustration of ground truth and inversion results of different methods on CVA.}
\end{figure}

\subsection{More Visualization Results on FFA and CFA}
\label{fig:fault_sup}
\begin{figure}[h]
    \centering
    \includegraphics[width=0.78\linewidth]{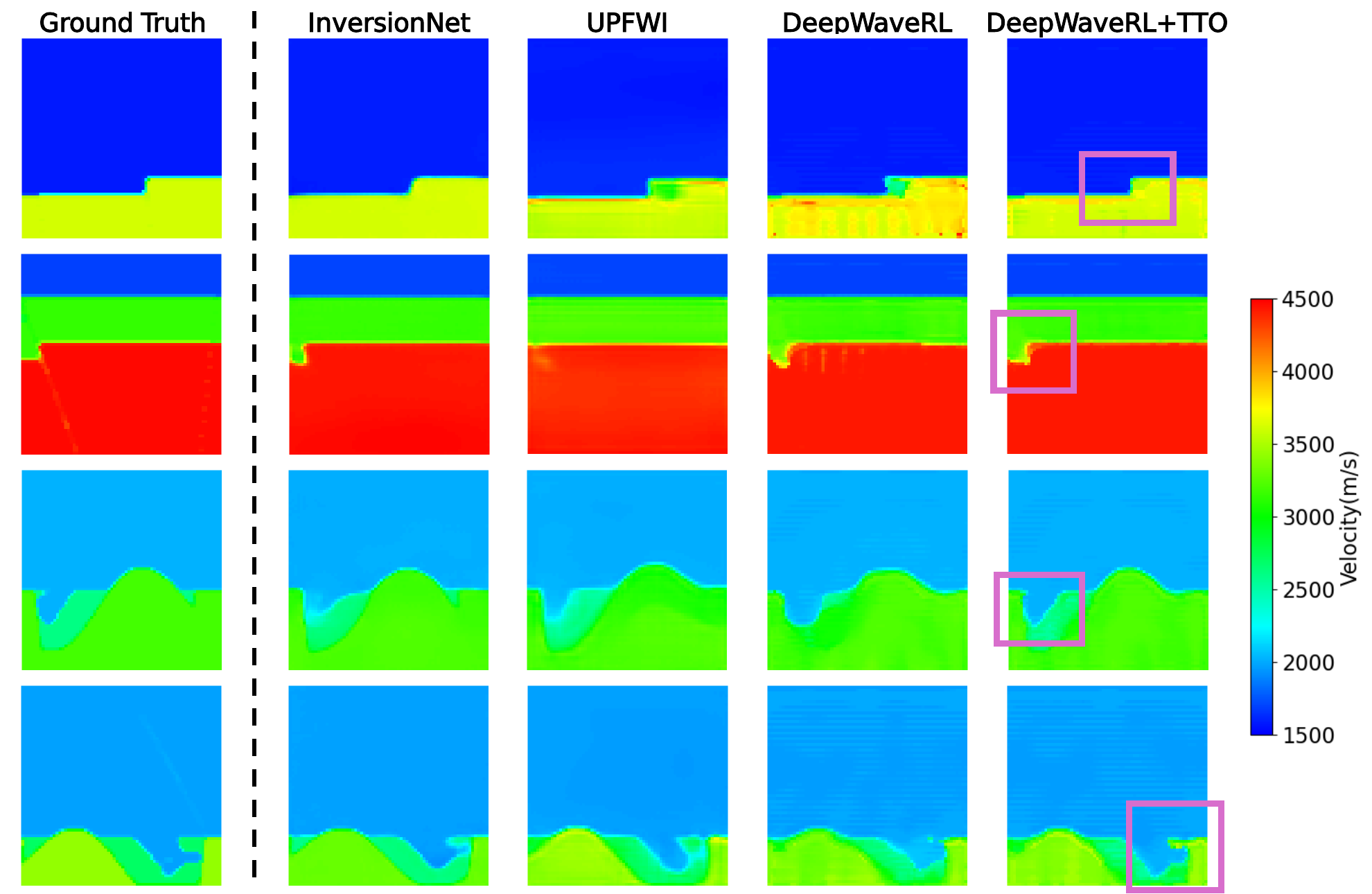}
    \caption{Illustration of ground truth and inversion results of different methods on FFA and CFA.}
\end{figure}
\clearpage

\subsection{Visualization Results of DeepWaveRL with Various Numbers of Bins}
\label{fig:bins}
\begin{figure}[h]
    \centering
    \includegraphics[width=0.78\linewidth]{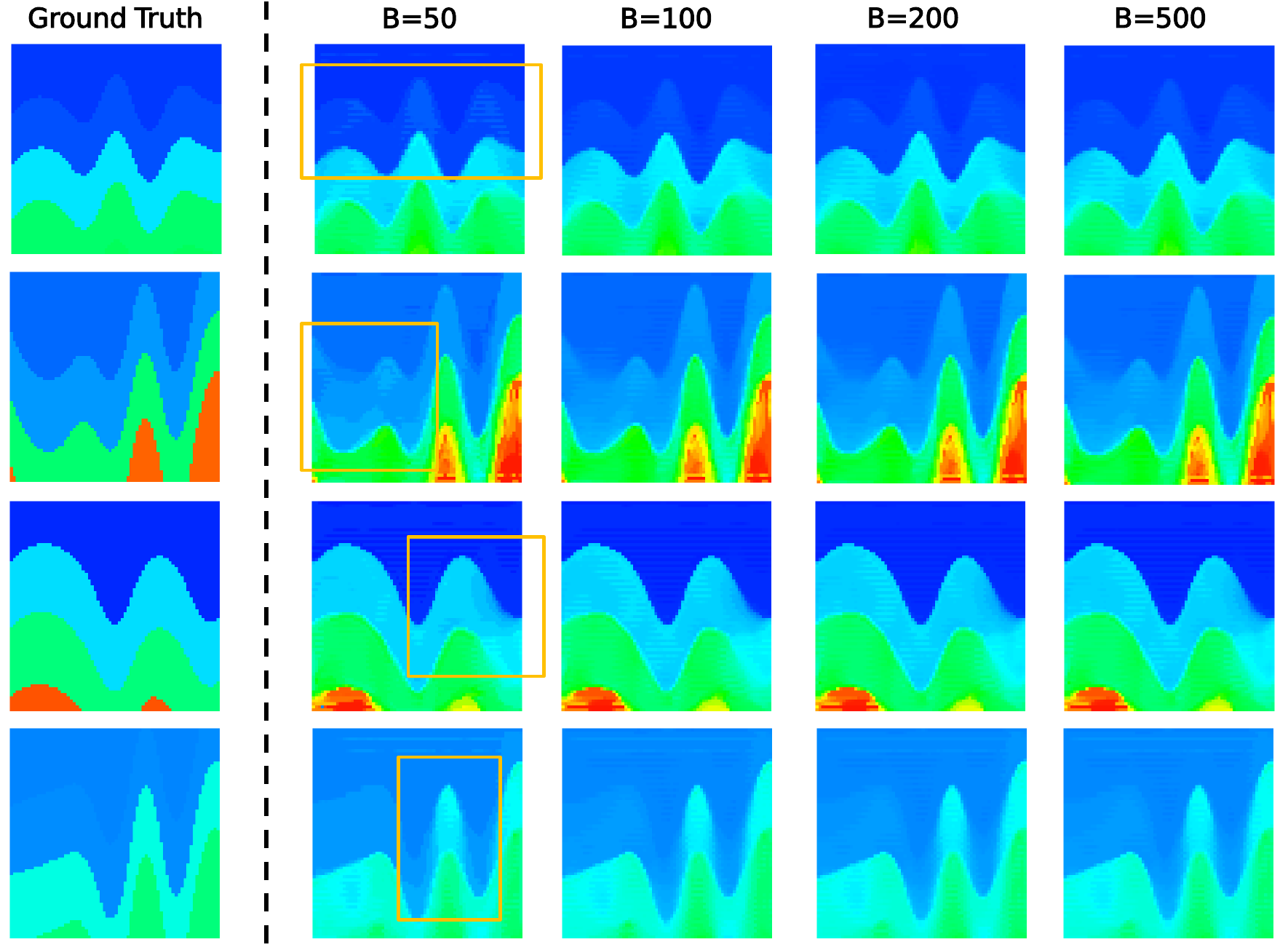}
    \caption{Comparison of DeepWaveRL+TTO with different numbers of bins on CVA. }
\end{figure}

\subsection{Visualization Results of DeepWaveRL without Logarithm Transformation}
\label{fig:wo_log}
\begin{figure}[h]
    \centering
    \includegraphics[width=0.9\linewidth]{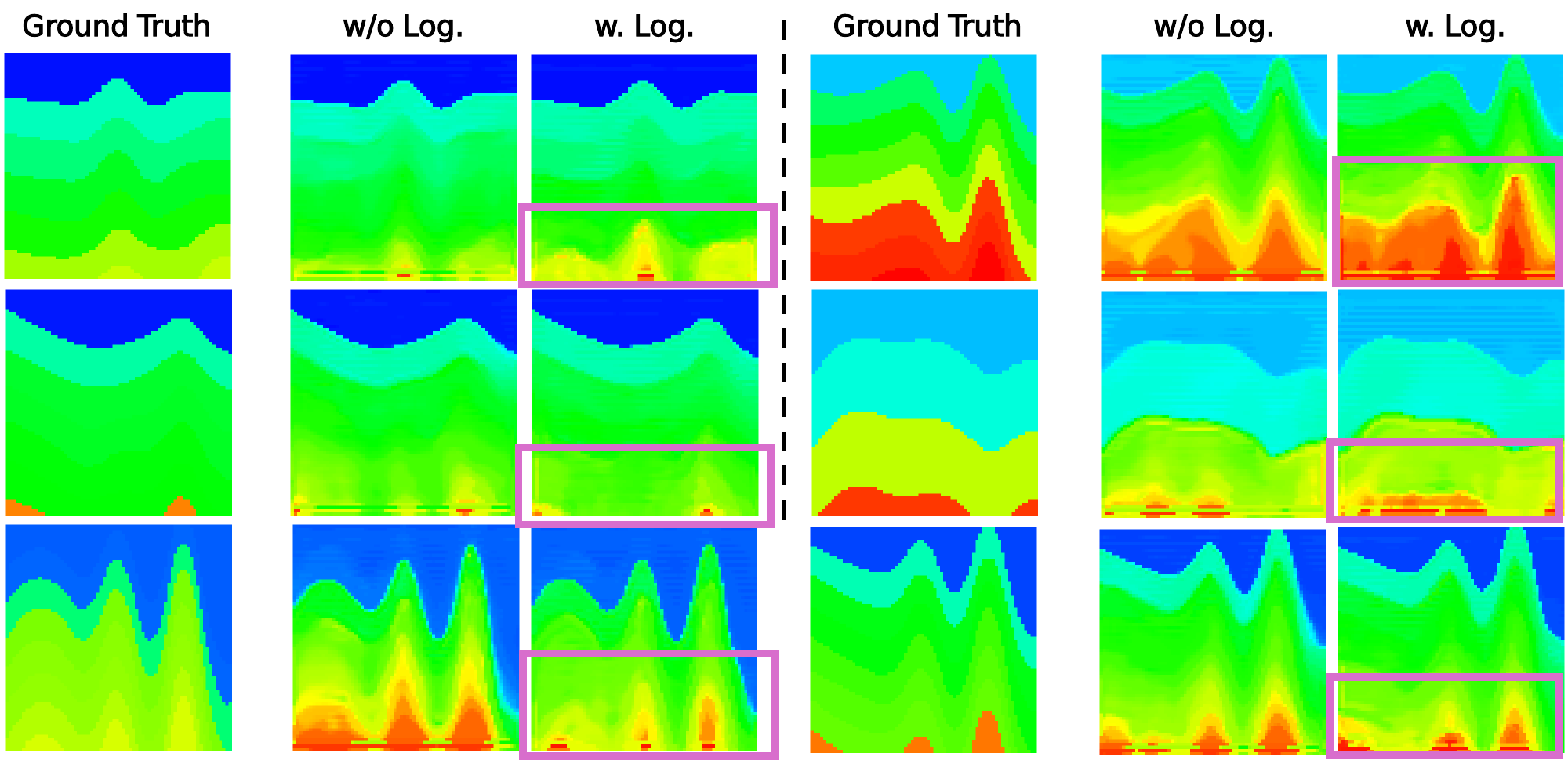}
    \caption{Illustration of ground truth and inversion results of DeepWaveRL with and without the sign-preserving logarithm transformation}
\end{figure}
\clearpage

\subsection{Visualization Results of DeepWaveRL with Various Group Sizes}
\label{fig:group}
\begin{figure}[h]
    \centering
    \includegraphics[width=0.9\linewidth]{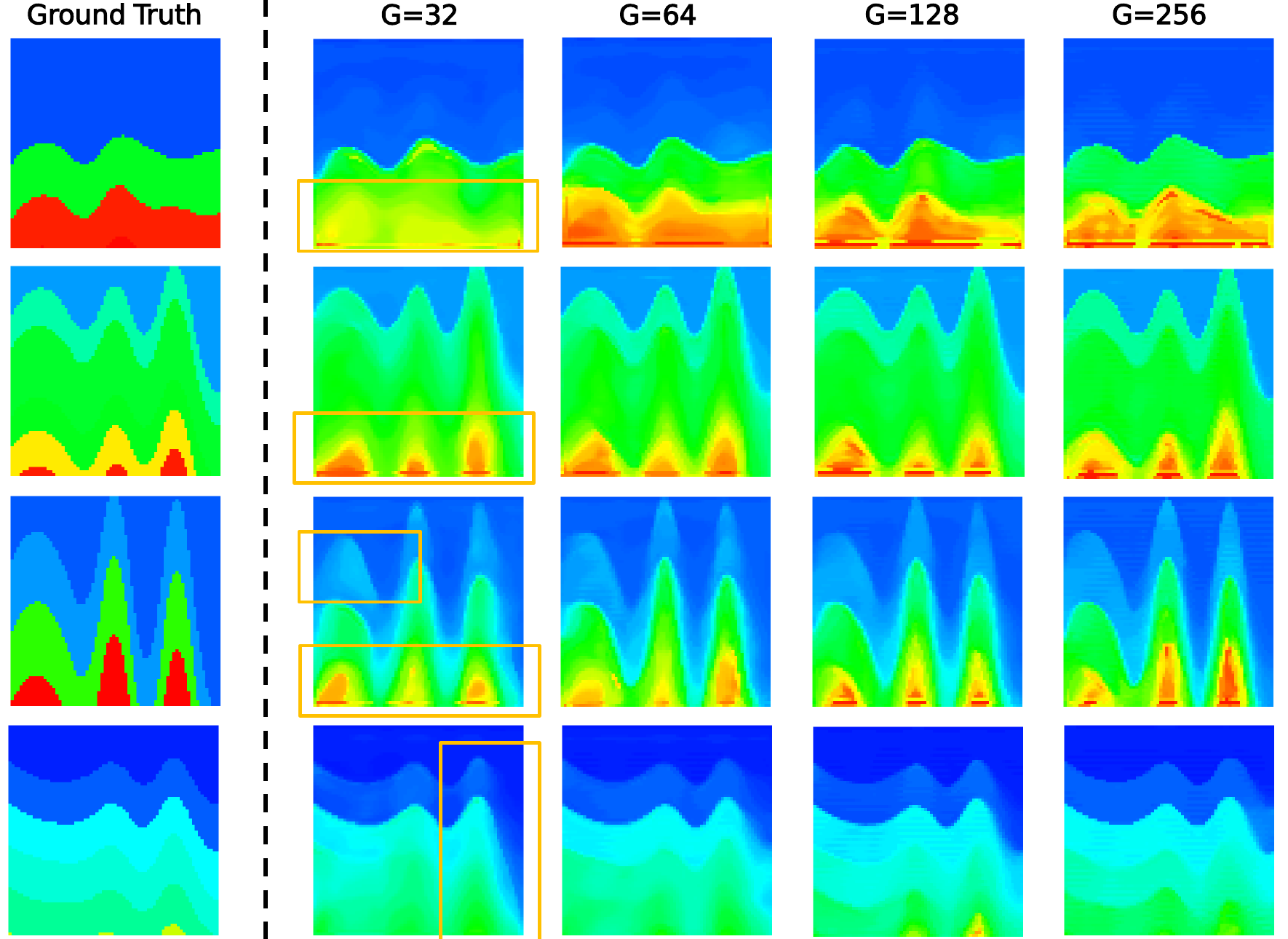}
    \caption{Comparison of DeepWaveRL with different group sizes on CVA.}
\end{figure}

\subsection{Visualization Results of DeepWaveRL without Refining the Prior}
\label{fig:tto_direct}
\begin{figure}[h]
    \centering
    \includegraphics[width=0.78\linewidth]{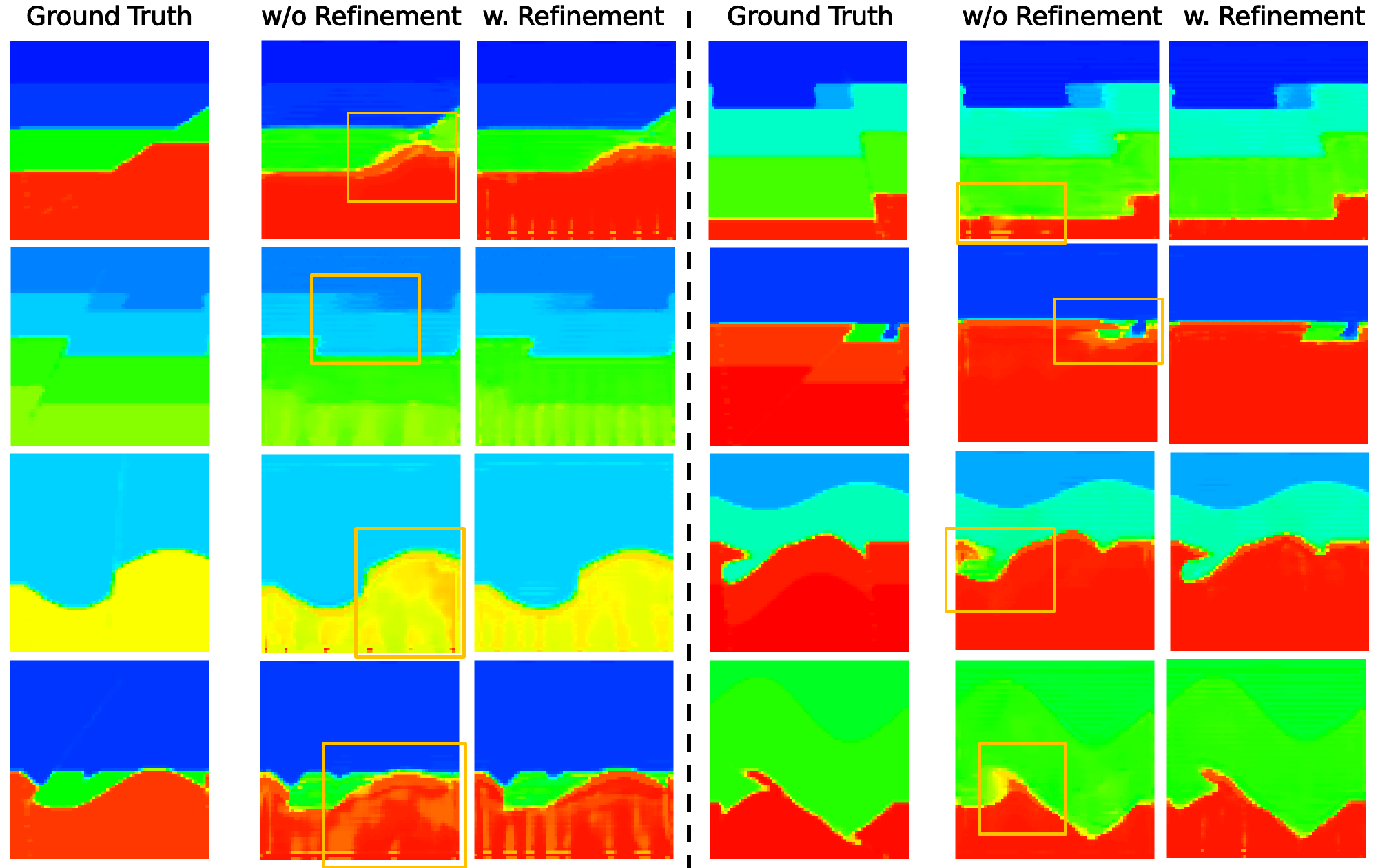}
    \caption{Comparison of DeepWaveRL+TTO with and without refining the prior on the training set of FFA (top two rows) and CFA (bottom two rows). }
\end{figure}

\clearpage
\subsection{Failure Cases}
\label{fig:fail}
\begin{figure}[h]
    \centering
    \includegraphics[width=0.65\linewidth]{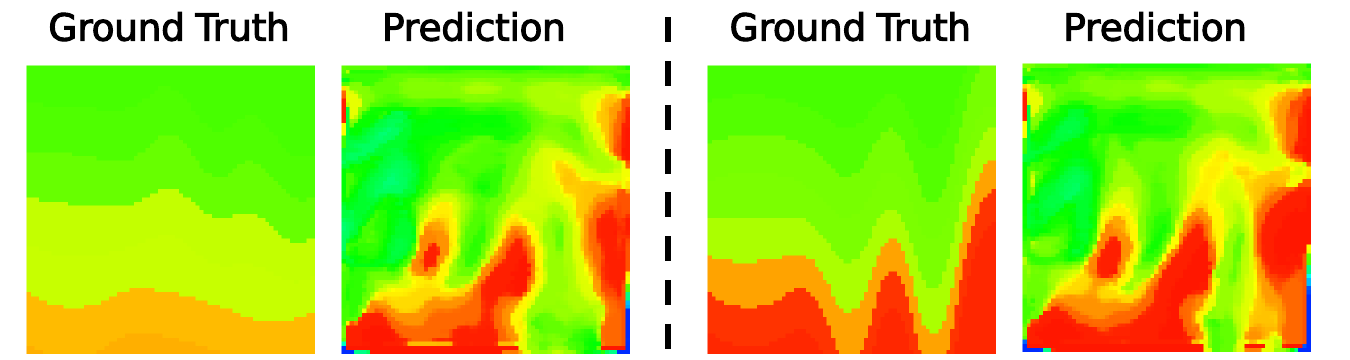}
    \caption{Examples of failure cases where the predictions collapse to similar patterns.}
\end{figure}

\subsection{Examples of Unstable Training Results}
\label{fig:unstable}
\begin{figure}[h]
    \centering
    \includegraphics[width=0.65\linewidth]{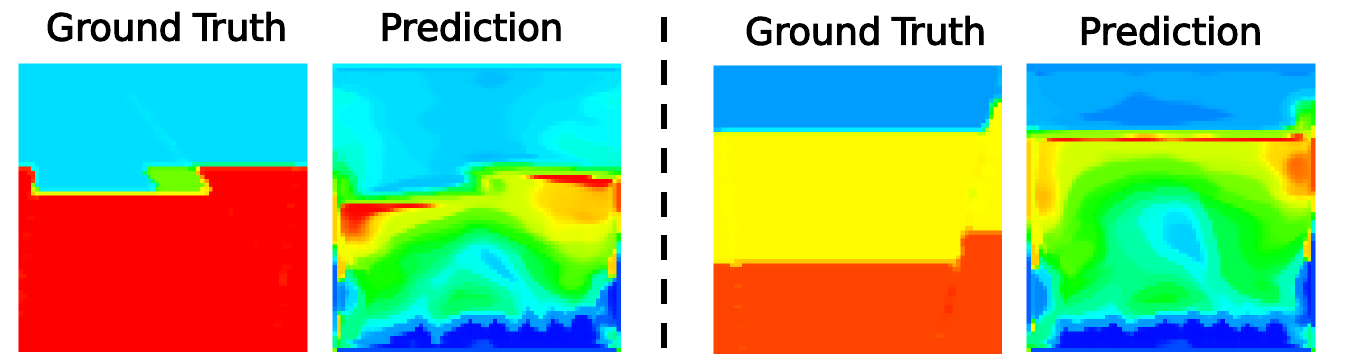}
    \caption{Examples of predicted velocity maps generated by the model that has experienced unstable training.}
\end{figure}


\end{document}